\RequirePackage{fix-cm}
\documentclass[smallextended]{svjour3}       
\smartqed  
\usepackage{text_style}
\usepackage{epsfig}
\usepackage{algorithm}

\usepackage{graphicx}
\usepackage{booktabs}
\usepackage{natbib}
\usepackage{mathtools}
\usepackage{times}
\usepackage{amsmath}
\usepackage{amssymb}
\usepackage{subcaption}
\usepackage{multirow}
\usepackage{siunitx}
\usepackage{algpseudocode}
\usepackage{hyperref}
\usepackage[title]{appendix}
\hypersetup{
  colorlinks, linkcolor=blue
}

\begin{document}

\title{Multimodal Image Synthesis with Conditional Implicit Maximum Likelihood Estimation
}


\author{Ke Li$^{\ast}$         \and Shichong Peng$^{\ast}$ \and Tianhao Zhang$^{\ast}$ \and Jitendra Malik
}


\institute{
Ke Li\at
             University of California, Berkeley \\
              \email{ke.li@eecs.berkeley.edu}           
           \and
           Shichong Peng \at University of Toronto \\
           \email{shichong.peng@mail.utoronto.ca}
           \and
           Tianhao Zhang \at Nanjing University \\
           \email{bryanzhang@smail.nju.edu.cn}
           \and
           Jitendra Malik \at University of California, Berkeley \\
           \email{malik@eecs.berkeley.edu} \\
           \\
$^{\ast}$ denotes equal contribution. 
}

\date{Received: date / Accepted: date}

\maketitle

\begin{abstract}
Many tasks in computer vision and graphics fall within the framework of conditional image synthesis. In recent years, generative adversarial nets (GANs) have delivered impressive advances in quality of synthesized images. However, it remains a challenge to generate both diverse and plausible images for the \emph{same} input, due to the problem of mode collapse. In this paper, we develop a new generic multimodal conditional image synthesis method based on Implicit Maximum Likelihood Estimation (IMLE) and demonstrate improved multimodal image synthesis performance on two tasks, single image super-resolution and image synthesis from scene layouts. We make our implementation publicly available.\footnote{Code for super-resolution is available at \url{https://github.com/niopeng/SRIM-pytorch} and code for image synthesis from scene layout is available at \url{https://github.com/zth667/Diverse-Image-Synthesis-from-Semantic-Layout}.}

\keywords{Conditional image synthesis \and Multimodal image synthesis \and Deep generative models \and Implicit maximum likelihood estimation}
\end{abstract}

\section{Introduction}
\label{intro}
In conditional image synthesis, the goal is to generate an image from some input, which can influence the image that is generated. It encompasses a broad range of tasks; examples include super-resolution, which aims to generate high-resolution images from low-resolution inputs, and image synthesis from scene layout, which aims to generate images from semantic segmentation maps. 

Deep learning has increasingly been used for image synthesis in recent years. Deep generative models, such as generative adversarial nets (GANs)~\citep{goodfellow2014generative,gutmann2014likelihood}, have emerged as one of the most popular approaches and have delivered impressive advances in image quality. Predominant approaches focus on the setting of generating a single image for each input image, which we will refer to as the \emph{unimodal prediction} problem. Relatively less attention has been devoted to the more general and challenging problem of \emph{multimodal prediction}, which aims to generate multiple equally plausible images for the \emph{same} input image. 

Why is the latter important? Conditional image synthesis is, by its very nature, ill-posed. That is, the information in the input is not enough to fully constrain the degrees of freedom in the output, and there are many plausible outputs that could all be consistent with the input. Therefore, we would like our system to be capable of generating all plausible outputs, rather than just selecting some plausible output arbitrarily. This could be important for downstream applications; for example, we may need to know the uncertainty of our system to estimate the informativeness of the input, or filter out a subset of the generated images to conform to some user-specified constraint. 

Extending GAN-based approaches to perform multimodal prediction has proven to be challenging, due to the problem of \emph{mode collapse}. More specifically, there is only one ground truth output for every input and the GAN objective encourages \emph{every} generated sample based on an input to be similar to the corresponding ground truth output. As a result, the generator tends to produce almost identical samples for the same input, regardless of the noise vector that is fed in~\citep{isola2017image}. This phenomenon is known as mode collapse, because the distribution of generated samples in this case collapses to a point mass on a mode of the underlying data distribution. 

\begin{figure}
    \centering
    \begin{subfigure}{0.13\textwidth}
    \includegraphics[width=\textwidth]{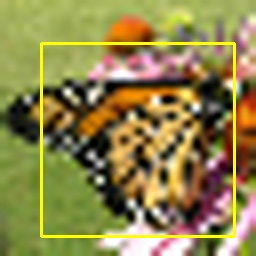}
    \caption{Input}
    \end{subfigure}
    \begin{subfigure}{0.77\textwidth}
    \includegraphics[width=\textwidth]{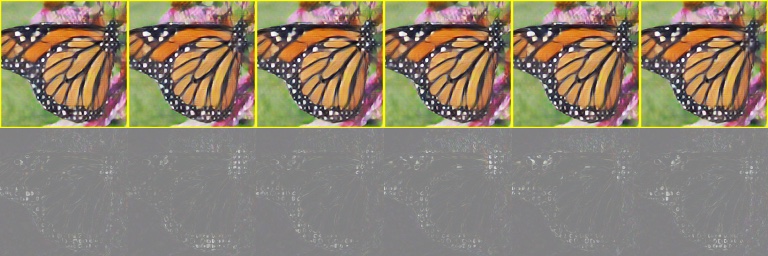}
    \caption{Samples}
    \end{subfigure}
    \caption{Samples generated by the proposed method (known as super-resolution implicit model, or SRIM for short) for the task of single image super-resolution (by a factor of 8). The top row shows different samples generated by our method, and the bottom row shows the difference between adjacent samples. As shown by the difference between the samples, the proposed method is able to generate diverse samples. See \url{https://people.eecs.berkeley.edu/~ke.li/papers/imle_img_synth/fig1.gif} for a visualization of different samples. }
    \label{fig:srim_intro}
\end{figure}

\begin{figure}
    \centering
    \begin{subfigure}{0.24\textwidth}
    \includegraphics[width=\textwidth]{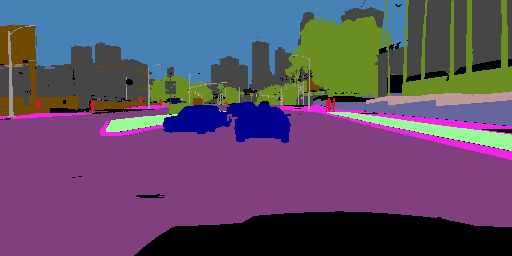}
    \caption{Input}
    \end{subfigure}
    \begin{subfigure}{0.7\textwidth}
    \includegraphics[width=\textwidth]{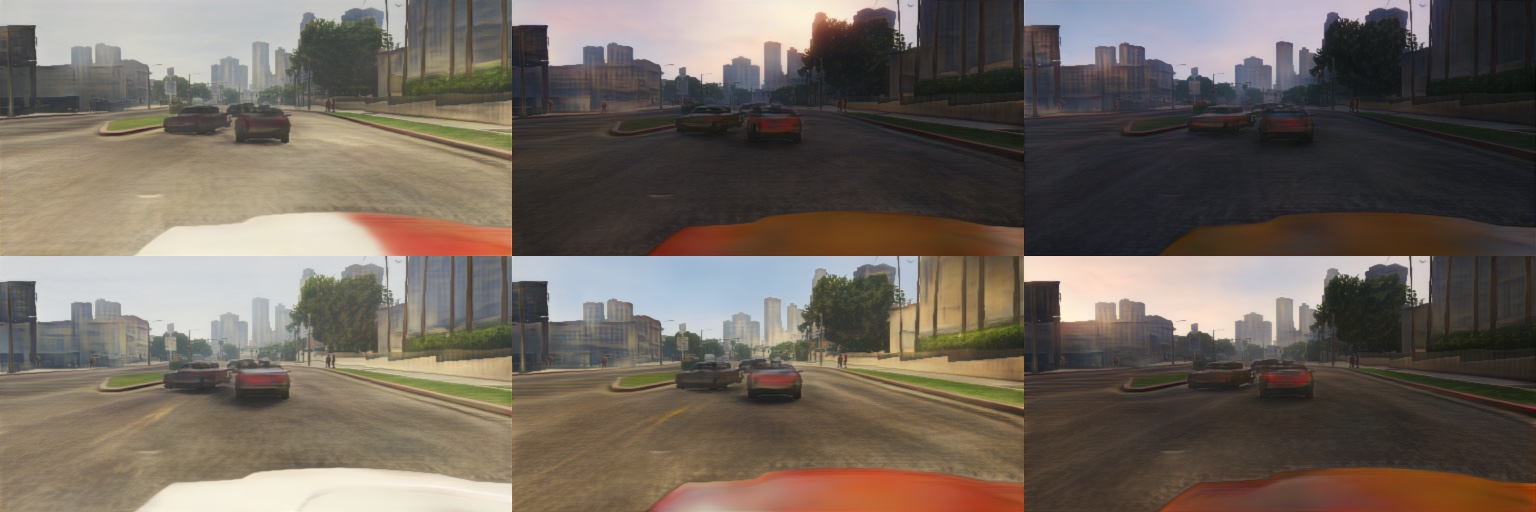}
    \caption{Samples}
    \end{subfigure}
    \caption{Samples generated by the proposed method for the task of image synthesis from scene layout. The group of images on the right are the different samples generated by our method. See \url{https://people.eecs.berkeley.edu/~ke.li/papers/imle_img_synth/fig2.gif} for a visualization of different samples.}
    \label{fig:imle_intro}
\end{figure}

Intuitively, this problem occurs because \emph{every} sample is made similar to the ground truth. This is undesirable, because there could be other images different from the ground truth that are also perfectly valid, due to the ill-posed nature of conditional image synthesis. Yet, generating any of such images would be penalized by the GAN objective; as a result, diversity is discouraged and mode collapse happens as a consequence. 

\begin{figure*}
    \centering
    \begin{subfigure}[t]{0.309\textwidth}
        \centering
        \includegraphics[width=\textwidth]{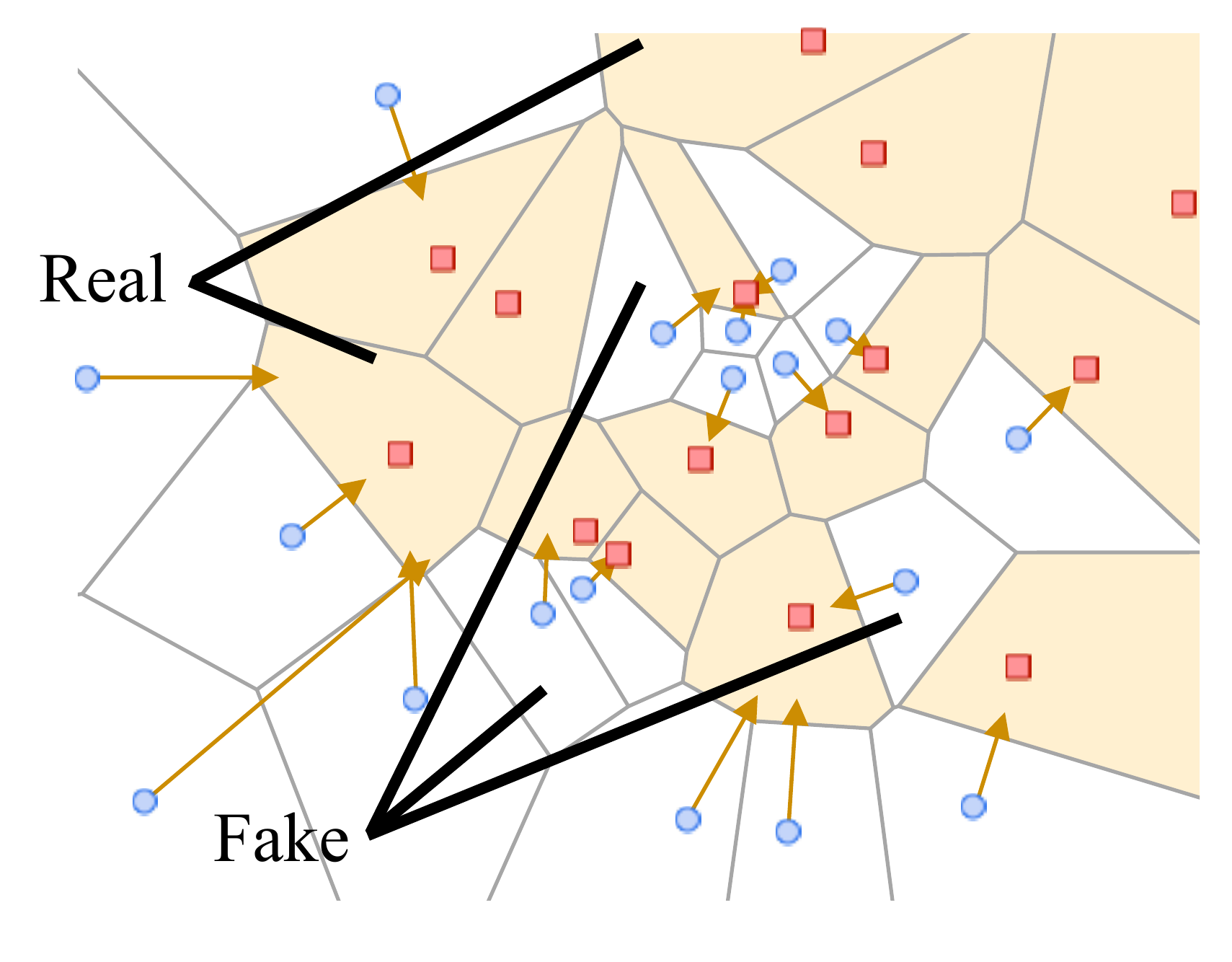}
        \caption{GAN\protect\\(Step 1)}\label{fig:schematic_gan1}
    \end{subfigure}
    \begin{subfigure}[t]{0.373\textwidth}
        \centering
        \includegraphics[width=\textwidth]{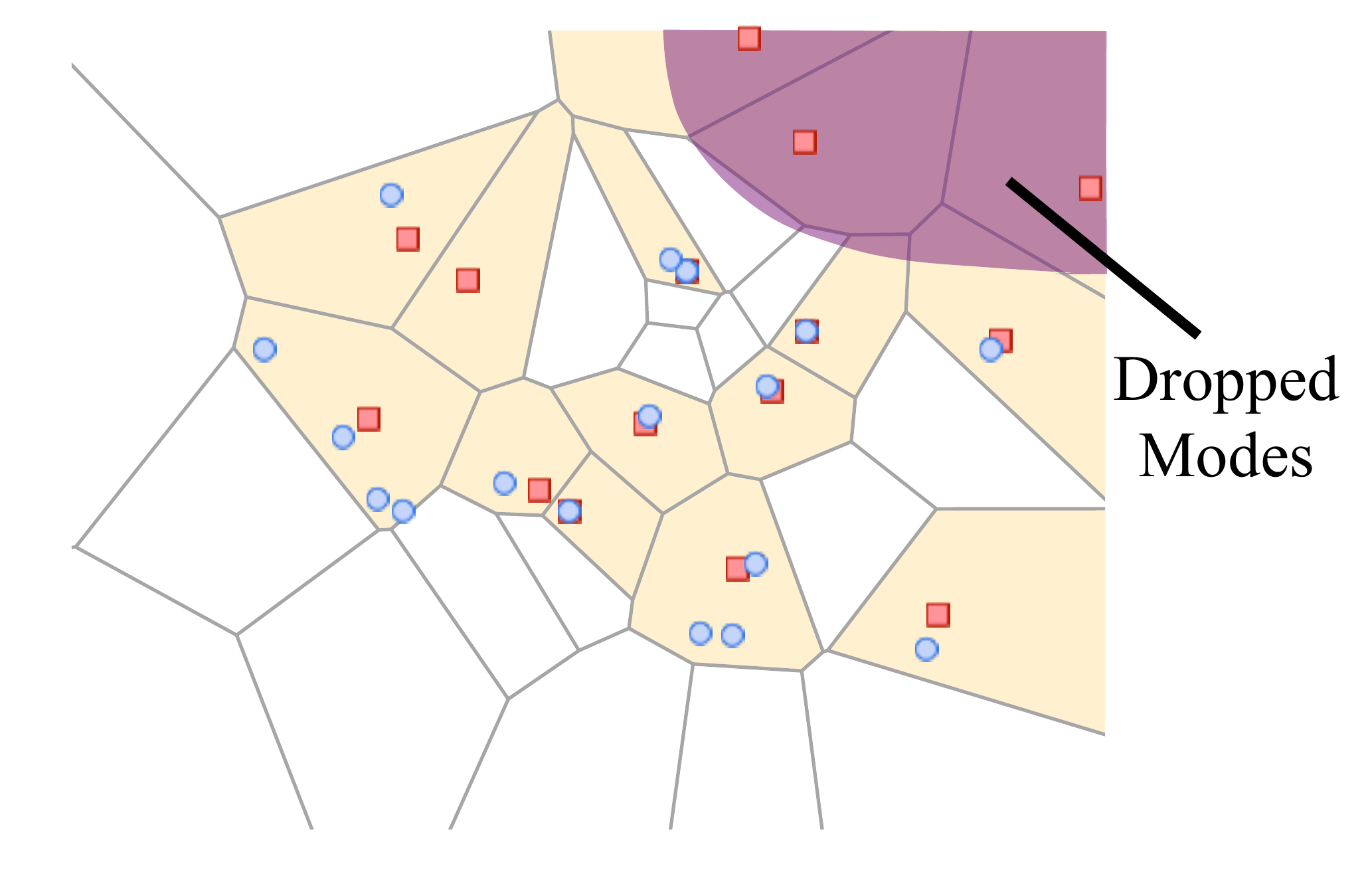}
        \caption{GAN\protect\\(Step 2)}\label{fig:schematic_gan2}
    \end{subfigure}
    \begin{subfigure}[t]{0.298\textwidth}
        \centering
        \includegraphics[width=\textwidth]{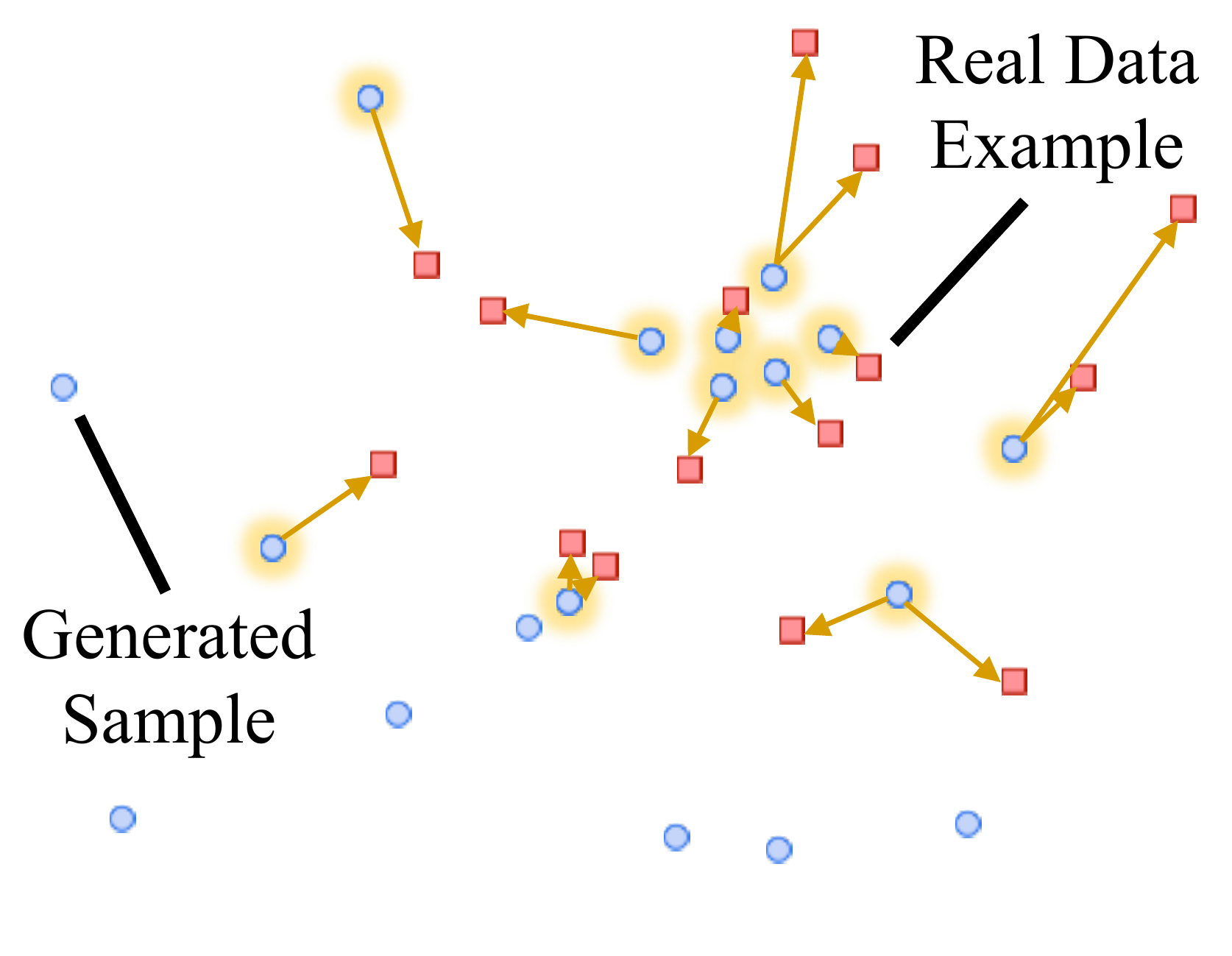}
        \caption{IMLE}\label{fig:schematic_imle}
    \end{subfigure}
    \caption{\label{fig:schematic}(a-b) How a (unconditional) GAN collapses modes (here we show a GAN with 1-nearest neighbour discriminator for simplicity). The blue circles represent generated images and the red squares represent real images. The yellow regions represent those classified as real by the discriminator, whereas the white regions represent those classified as fake. As shown, when training the generator, each \emph{generated image} is essentially pushed towards the nearest \emph{real image}. Some real images may not be selected by any generated image during training and therefore could be ignored by the trained generator -- this is a manifestation of mode collapse. (c) An illustration of how Implicit Maximum Likelihood Estimation (IMLE) works. IMLE avoids mode collapse by reversing the direction in which generated images are matched to real images. Instead of pushing each generated image towards the nearest real image, for every real image, it pulls the nearest generated image towards it -- this ensures that all real images are matched to some generated image, and no real images are ignored.}
    
\end{figure*}

We propose a different approach, based on the method of Implicit Maximum Likelihood Estimation~\citep{li2018implicit}. At a high level, IMLE is similar in principle to GANs in that it tries to make generated samples similar to real data examples, but it has one important difference: it swaps \emph{the direction} in which generated samples are made similar to real data examples. More concretely, the GAN objective encourages \emph{each} generated sample to be similar to \emph{some} real data example. On the other hand, IMLE encourages \emph{each} real data example to be similar to \emph{some} generated sample. This difference is illustrated schematically in Figure~\ref{fig:schematic}.

The original IMLE method was developed for the unconditional setting. In this paper, we develop an extension of IMLE to the conditional setting, where the generative model takes in an image as input (in addition to the latent noise vector). There is a simple intuitive interpretation of the difference between conditional IMLE and conditional GANs: for a given input image, in a conditional GAN, \emph{every} sample is made similar to the ground truth, whereas in conditional IMLE, only \emph{one} of the samples is made similar to the ground truth. 

A natural question arises: why wouldn't conditional IMLE generate spurious implausible samples if only \emph{one} of the samples is made similar to the ground truth? This is because the optimization is performed with respect to the \emph{parameters} of the generative model, rather than the samples themselves. Since the probability density associated with the model always integrates to 1, if the density near the ground truth is made higher, density must be taken away from other regions, which are precisely the regions containing implausible samples. 

\section{Related Work}
\label{sec:related_work}

\subsection{Unimodal Prediction}
Many image synthesis methods developed in recent years are based on generative adversarial nets (GANs)~\citep{goodfellow2014generative,gutmann2014likelihood}. Most of these methods are capable of producing only a single image for each given input, due to the problem of mode collapse. Various work has explored conditioning on different types of information. Some methods condition on a scalar, such as object category and attribute~\citep{mirza2014conditional,gauthier2014conditional,denton2015deep}. Other methods condition on richer labels, such as text description~\citep{reed2016learning}, surface normal maps~\citep{wang2016generative}, previous frames in a video~\citep{mathieu2015deep,vondrick2016generating} and images~\citep{yoo2016pixel,isola2017image,zhu2017unpaired}. Some methods only condition on inputs images in the generator, but not in the discriminator~\citep{pathak2016context,ledig2017photo,zhu2016generative,li2016precomputed}. \citet{kaneko2017generative}, \citet{reed2016learning} and \citet{sangkloy2017scribbler} explore conditioning on attributes that can be modified manually by the user at test time; these methods are not true multimodal methods because they require manual changes to the input (rather than just sampling from a fixed distribution) to generate a different image. 

Another common approach to image synthesis is to treat it as a simple regression problem. To ensure high perceptual quality, the loss is usually defined on some fixed transformation of the raw pixels, e.g.: features from some intermediate layers of a pretrained neural net. This paradigm has been applied to super-resolution \citep{bruna2015super,johnson2016perceptual}, style transfer \citep{johnson2016perceptual} and video frame prediction \citep{srivastava2015unsupervised,oh2015action,finn2016unsupervised}. These methods are by design unimodal methods because neural nets are functions, and so for the same input image, the neural net must synthesize the same output image. 

Various methods have been developed for the problem of image synthesis from semantic layouts. For example, \citet{karacan2016learning} developed a conditional GAN-based model for generating images from semantic layouts and labelled image attributes. It is important to note that the method requires supervision on the image attributes and is therefore a unimodal method. (Multimodal methods must be able to discover the different modes in an unsupervised fashion, because if mode labels were given, the synthesis problem would become equivalent to modelling the distribution over images conditioned on \emph{both} the input image and the mode label. Note that if this distribution were multimodal, which could happen if the mode label does not uniquely identify a mode, then the different modes of this distribution must still be learned without supervision. Methods that cannot do this are not true multimodal methods.) \citet{isola2017image} developed a conditional GAN that can generate images solely from semantic layout. However, it is only able to generate a single plausible image for each semantic layout, due to the problem of mode collapse in GANs. \citet{wang2017high} further refined the approach of \citet{isola2017image}, focusing on the high-resolution setting. While these methods are able to generate images of high visual fidelity, they are all unimodal methods. 

\subsection{Fixed Number of Modes}

A simple approach to generate a fixed number of different outputs for the same input is to use different branches or models for each desired output. For example, \citet{guzman2012multiple} proposed a model that outputs a fixed number of different predictions simultaneously, which was an approach adopted by \citet{chen2017photographic} to generate different images for the same semantic layout. Unlike most approaches, \citet{chen2017photographic} did not use the GAN framework; instead it uses a simple feedforward convolutional network. On the other hand, \citet{ghosh2017multi} uses a GAN framework, where multiple generators are introduced, each of which generates a different mode. The above methods all have two limitations: (1) they are only able to generate a fixed number of images for the same input, and (2) they cannot generate continuous changes. 

\subsection{Arbitrary Number of Modes}
\label{sec:related_word_arb}

A number of GAN-based approaches propose adding learned regularizers that discourage mode collapse. BiGAN/ALI~\citep{donahue2016adversarial,dumoulin2016adversarially} trains a model to reconstruct the latent code from the image; however, when applied to the conditional setting, significant mode collapse still occurs because the encoder is not trained until optimality and so cannot perfectly invert the generator. VAE-GAN~\citep{larsen2015autoencoding} combines a GAN with a variational autoencoder (VAE), in order to take advantage of the lack of mode collapse in VAEs. However, image quality suffers because the generator is trained on latent code sampled from a mixture of the encoder/approximate posterior and the prior. At test time, only the prior is available, so the latent code is sampled from just the prior, resulting in a mismatch between training and test conditions. \citet{zhu2017toward} proposed Bicycle-GAN, which combines both of the above approaches. While this alleviates the above issues, it is difficult to train, because it requires training three different neural nets simultaneously, namely the generator, the discriminator and the encoder. Because they serve opposing roles and effectively regularize one another, it is important to strike just the right balance, which makes it hard to train successfully in practice. 

A related but different line of work is on unpaired image-to-image translation~\citep{huang2018multimodal, Ma2018ExemplarGU, Lee2018DRITDI, Lee2019HarmonizingML,  Almahairi2018AugmentedCL, Yang2019DiversitySensitiveCG}. \citep{huang2018multimodal} takes the content code and combine with arbitrary style code in the target domain. \citep{Ma2018ExemplarGU} used exemplar guidance from the target domain and feature masking for semantic consistency. \citep{Lee2018DRITDI} adopted cross-cycle consistency loss based on disentangled representations. Most papers make the simplifying assumption that the factors of variation in images in the target domain are similar across different corresponding images in the source domain, and so they can decompose the representation of an image into the Cartesian product of domain-invariant content code and domain-specific style code. As a concrete example, for the task of image synthesis from scene layout, the possible factors of variation in the output image are similar across different scene layouts, i.e.: changes in weather conditions or times of the day, and so this assumption holds and these methods are well-suited to this task. However, on other tasks, the possible factors of variation in the output image depend on what the input image is and vary significantly for different input images. For example, for the task of super-resolution, the factors of variation are the high-frequency details that do not exist in the low-resolution input; the kinds of details that are plausible are highly dependent on the content of the input image. As a result, Cartesian decomposition could give rise to spurious samples, and so the above methods are ill-suited to this task. 

A number of methods for colourization~\citep{charpiat2008automatic,zhang2016colorful,larsson2016learning} predict a discretized marginal distribution over colours of each individual pixel. Similarly, approaches based on PixelCNN, e.g.:~\citep{Dahl2017PixelRS}, discretize the distribution over the colour of a pixel conditioned on other pixels. While these approaches are able to capture multimodality in the marginal or conditional distribution, ensuring global consistency between different parts of the image is not easy, since there are correlations between the colours of different pixels. It is difficult to learn such correlations because these approaches do not directly learn the joint distribution over the colours of \emph{all} pixels. 

\section{Method}
\label{sec:method}
\subsection{Background}

In conditional image synthesis, the goal is to generate images that are consistent with the input. In general, this problem is ill-posed, since there could be many different plausible high-resolution images that are all consistent with the input. The problem of \emph{multimodal} conditional image synthesis requires generating \emph{all} such images. This contrasts with the simpler problem of \emph{unimodal} conditional image synthesis, which only requires generating \emph{one} such image. 

This can be naturally formulated as a probabilistic modelling problem. Let $\mathbf{x}$ denote a random variable that corresponds to the input image and $\widetilde{\mathbf{y}}$ denote a random variable that corresponds to the output image. The distribution we would like to model is therefore the conditional distribution of $\widetilde{\mathbf{y}}$ given $\mathbf{x}$, i.e.: $p(\widetilde{\mathbf{y}} | \mathbf{x})$. Each plausible output image that is consistent with the input image would be a mode of the distribution; because there could be many plausible images that are consistent with the same input image, $p(\widetilde{\mathbf{y}} | \mathbf{x})$ is usually multimodal. Different output images could be then generated by sampling from our model of this distribution. 

We will make the dependence of the model distribution on the model parameters $\theta$ explicit by writing $p(\widetilde{\mathbf{y}} | \mathbf{x}; \theta)$. To learn $\theta$, we would like to maximize the log-likelihood of the ground truth outputs given the corresponding inputs. That is, given a training dataset $\mathcal{D}=\left\{(\mathbf{x}_1,\mathbf{y}_1),\ldots,(\mathbf{x}_n,\mathbf{y}_n)\right\}$, we would like to solve the following optimization problem:

\[
\hat{\theta}_{\mathrm{MLE}} \coloneqq \arg\max_{\theta} \sum_{i=1}^{n} \log p(\mathbf{y}_i | \mathbf{x}_i;\theta)
\]

\subsection{Probabilistic Model}
\label{sec:prob_model}
We will model $p(\widetilde{\mathbf{y}} | \mathbf{x})$ using an implicit probabilistic model, that is, a probabilistic model that is defined in terms of a sampling procedure. This contrasts with classical probabilistic models (also known as prescribed probabilistic models) like Markov Random Fields (MRFs), which are defined by explicitly specifying the form of the probability density. One example of an implicit probabilistic model $p(\widetilde{\mathbf{y}} | \mathbf{x};\theta)$ is the generator in a (conditional) GAN, which is defined by the following sampling procedure:

\begin{enumerate}
\item Sample $\mathbf{z} \sim \mathcal{N}(0,\mathbf{I})$
\item Return $\widetilde{\mathbf{y}} \coloneqq T_{\theta}(\mathbf{x},\mathbf{z})$
\end{enumerate}

Here, $T_{\theta}$ is a deep convolutional neural net that takes in two inputs, the input $\mathbf{x}$ and the latent noise vector $\mathbf{z}$. Compared to classical (prescribed) probabilistic models, such implicit probabilistic models have an advantage in that they can be made more expressive without complicating inference. However, this comes at the cost of making training more difficult, because the density and therefore the log-likelihood cannot in general be computed tractably. This necessitates the use of a likelihood-free method for training/parameter estimation. 

\subsection{Training Method}

One example of a likelihood-free method is the GAN training procedure, which works by introducing a discriminator and making all the samples indistinguishable (to the discriminator) from the real data. However, due to the well-documented issue of mode collapse~\citep{arora2017gans,isola2017image}, training an implicit model this way results in the the latent input $\mathbf{z}$ being ignored. As a result, for the same input image, only one output image can be generated. This is therefore unsuitable for the task of multimodal image synthesis. 

Recently, another likelihood-free method known as Implicit Maximum Likelihood Estimation (IMLE) was introduced~\citep{li2018implicit}, which does not suffer from mode collapse. Whereas GANs cannot be made equivalent to maximum likelihood~\citep{goodfellow2014distinguishability} (in fact they have been shown to be roughly equivalent to minimizing the reverse KL divergence, which is the opposite of maximum likelihood in the sense that the arguments of the KL divergence are reversed compared to maximum likelihood~\citep{arjovsky2017towards}), IMLE can be shown to be equivalent to maximum likelihood (under appropriate conditions). This has practical significance -- maximum likelihood prohibits mode collapse because when the generator collapses modes, by definition, it is unable to generate images at the collapsed modes. Then the probability density that the generator assigns to images at the collapsed modes must be zero or near-zero, and assuming those modes are observed in the training data, the generator must assign near-zero density to some training images. Because the likelihood is the product of probability densities at different training images, having a near-zero density at any of the training images will result in near-zero likelihood, which most definitely will not maximize likelihood. 

\subsection{Implicit Maximum Likelihood Estimation}

The IMLE method was designed for the unconditional setting, where the goal is to train an implicit model for the marginal distribution over images $p(\widetilde{\mathbf{y}})$. The implicit model $p(\widetilde{\mathbf{y}};\theta)$ in this setting is slightly different and is defined by the following sampling procedure:

\begin{enumerate}
\item Sample $\mathbf{z} \sim \mathcal{N}(0,\mathbf{I})$
\item Return $\widetilde{\mathbf{y}} \coloneqq T_{\theta}(\mathbf{z})$
\end{enumerate}

Notice that the neural net $T_{\theta}(\cdot)$ now takes in one input, the latent noise vector $\mathbf{z}$, as opposed to two inputs, as in the conditional setting described in Section~\ref{sec:prob_model}. IMLE finds a setting of the parameters of the neural net that minimizes the distance from each training example to the nearest generated sample among a pool of $m$ samples generated from the model. More precisely, given a set of training examples $\mathbf{y}_1,\ldots,\mathbf{y}_n$, IMLE optimizes the following training objective:
\begin{align*}
\hat{\theta}_{\mathrm{IMLE}} \coloneqq \;& \arg\min_{\theta}\mathbb{E}_{\mathbf{z}_{1},\ldots,\mathbf{z}_{m} \sim \mathcal{N}(0, \mathbf{I})}\left[\sum_{i=1}^{n}\min_{j\in \{1,\ldots,m\}}\mathcal{L}(T_{\theta}(\mathbf{z}_{j}), \mathbf{y}_{i})\right],
\end{align*} where $\mathcal{L}(\cdot,\cdot)$ is a distance metric and $m$ is a hyperparameter. 

Because IMLE encourages each training example to have a nearby generated sample, no training example can be ignored, thereby avoiding mode collapse. 

\subsection{Conditional IMLE}

In the conditional setting, the goal is to learn the distribution over output images \emph{conditioned} on the input image, $p(\widetilde{\mathbf{y}}  | \mathbf{x})$. In this paper, we propose an extension of IMLE that can train a \emph{conditional} implicit model $p(\widetilde{\mathbf{y}} | \mathbf{x};\theta)$, such as the one defined in Section~\ref{sec:prob_model}. 

The probabilistic model $p(\widetilde{\mathbf{y}} | \mathbf{x};\theta)$ now models a different distribution for every value of $\mathbf{x}$, which requires two changes to be made to IMLE. First, the value of $\mathbf{x}$ must be provided to the neural net $T_{\theta}$ in order to sample from the desired distribution. Second, the samples for different values of $\mathbf{x}$ must be kept separate since they are from different distributions. Consequently, for each input data example $\mathbf{x}_i$, we must look for the nearest sample among the samples generated from $p(\widetilde{\mathbf{y}} | \mathbf{x}_i)$ (and not $p(\widetilde{\mathbf{y}} | \mathbf{x}_j)$ where $j \neq i$). 

Concretely, given a set of training inputs and outputs $(\mathbf{x}_1,\mathbf{y}_1),\ldots,(\mathbf{x}_n,\mathbf{y}_n)$, conditional IMLE optimizes the following training objective:
\begin{align*}
\hat{\theta}_{\mathrm{IMLE}} \coloneqq \;& \arg\min_{\theta}\mathbb{E}_{\mathbf{z}_{1,1},\ldots,\mathbf{z}_{n,m} \sim \mathcal{N}(0, \mathbf{I})}\left[\sum_{i=1}^{n}\min_{j\in \{1,\ldots,m\}}\mathcal{L}(T_{\theta}(\mathbf{x}_{i},\mathbf{z}_{i,j}), \mathbf{y}_{i})\right],
\end{align*} where $\mathcal{L}(\cdot,\cdot)$ is a distance metric and $m$ is a hyperparameter. 

The training procedure is presented in Algorithm~\ref{alg:cimle}.

\begin{algorithm}
\caption{\label{alg:cimle}Conditional Implicit Maximum Likelihood Estimation (Conditional IMLE) Procedure}
\begin{algorithmic}
\Require The set of inputs and corresponding target outputs $\left\{ (\mathbf{x}_{i},\mathbf{y}_{i})\right\} _{i=1}^{n}$ and a distance metric $\mathcal{L}(\cdot,\cdot)$
\State Initialize the parameters $\theta$ of the neural net $T_\theta$
\For{$k = 1$ \textbf{to} $K$}
    \State Pick a random batch $S \subseteq \{1,\ldots,n\}$
    \For{$i \in S$}
        \State Randomly generate i.i.d. $m$ noise vectors $\mathbf{z}_{i,1},\ldots,\mathbf{z}_{i,m}$
        \State $\widetilde{\mathbf{y}}_{i, j} \gets T_{\theta}(\mathbf{x}_{i}, \mathbf{z}_{i,j})\; \forall j \in [m]$
        \State $\sigma(i) \gets \arg \min_{j} \mathcal{L}(\widetilde{\mathbf{y}}_{i, j},\mathbf{y}_{i})\;\forall j \in [m]$
    \EndFor
    \For{$l = 1$ \textbf{to} $L$}
    \State Pick a random mini-batch $\widetilde{S} \subseteq S$
    \State $\theta \gets \theta - \eta \nabla_{\theta}\left(\sum_{i \in \widetilde{S}}\mathcal{L}(\widetilde{\mathbf{y}}_{i, \sigma(i)}, \mathbf{y}_{i})\right) / |\widetilde{S}|$
    \EndFor
\EndFor
\State \Return $\theta$
\end{algorithmic}
\end{algorithm}

\subsection{Computational Efficiency}

Finding the nearest generated sample to a training example is a nearest neighbour search problem. When the dimensionality is high, as is usually the case with images, this is traditionally believed to be hard, because the curse of dimensionality renders most exact and approximate nearest neighbour search algorithms ineffective. This is no longer an issue, due to a new family of exact randomized nearest neighbour algorithms we developed known as Dynamic Continuous Indexing (DCI)~\citep{li2016fast,li2017fast}, which can overcome both the curse of ambient dimensionality and the curse of intrinsic dimensionality. We use DCI in our implementation, which makes it possible to find the nearest sample to each training example very computationally efficient, even when the dimensionality is in the thousands or tens of thousands. In fact, finding the nearest samples takes less than 1\% of the execution time of the entire algorithm. 

\subsection{Distance Metric}

We choose a distance metric $\mathcal{L}(\cdot,\cdot)$ that reflects perceptual similarity between images. To this end, we follow standard practice in the literature and use a weighted combination of distances on pixels and features at selected layers of a VGG-19 network~\citep{simonyan2014very} pretrained on ImageNet. More precisely, the distance between a synthesized image sampled from the model $\widetilde{\mathbf{y}}$ and a ground truth image $\mathbf{y}$ takes the following form:
\begin{equation}\label{eq:loss}
    \mathcal{L}(\widetilde{\mathbf{y}}, \mathbf{y}) = \sum_{i=1}^l\lambda_i\left\Vert\Phi_i(\widetilde{\mathbf{y}}) - \Phi_i(\mathbf{y})\right\Vert, 
\end{equation}
where $\Phi_i(\cdot)$ denotes the raw pixels or the activations/features at a layer of the pretrained VGG-19 network applied to the input to the function, $\lambda_i$ denotes the weight associated with the features of a layer, and $\left\Vert \cdot \right\Vert$ denotes a $\ell_p$-norm (usually $\ell_1$ and $\ell_2$ norm). The choice of norm and layers whose activations are used to compute distance are task-dependent. Specific choices for each task will be discussed in Section~\ref{seg:experiments}. The weights $\{\lambda_i\}_{i=1}^l$ are set so that each term in the distance metric is roughly of the same magnitude. 

\section{Tasks}
\label{seg:tasks}

We consider two different conditional image synthesis tasks in this paper, single image super-resolution and image synthesis from scene layout. We describe them below. 

\subsection{Single Image Super-Resolution}

Given a low-resolution input image, the problem of super-resolution requires the prediction of a high-resolution output image that is consistent with the input. This problem is ill-posed, since there could be many high-resolution images that are all plausible and consistent with the input image. We consider the multimodal version of this problem, where the objective is to predict \emph{all} plausible versions of the input image at high resolution. This is challenging, because in the training data, we only observe one high-resolution image for every low-resolution image and never observe alternative versions of high-resolution images that are consistent with the same low-resolution image. In contrast, most prior super-resolution methods consider the unimodal version of this problem, which only requires predicting a \emph{single} plausible version of the input image at high resolution. Also, we consider the setting of upscaling by a factor of 8, i.e.: where the width and height increase by a factor of 8. This is more challenging than the settings considered by most prior super-resolution methods, which are designed for upscaling factors of 4 or less. More precisely, given a low-resolution input image $\mathbf{x} \in [0,255]^{h\times w \times 3}$, our goal is to predict plausible high-resolution images $\widetilde{\mathbf{y}} \in [0,255]^{H\times W \times 3}$ that when downsampled are similar to the given low-resolution input image. In our case, $H = 8h$ and $W = 8w$. 

\subsection{Image Synthesis from Scene Layout}

Given a scene layout represented by a semantic segmentation map (which specifies the object category label of every pixel), the problem of image synthesis from scene layout requires the prediction of an image that is consistent with the provided semantic segmentation map. This problem is ill-posed, because a semantic segmentation map does not specify the appearance of the objects in the image, and so there is no single image that can be recovered from the semantic segmentation. Instead, many images are both plausible and consistent with the input semantic segmentation map; for example, the images could all have identical scene layouts but differ in the illumination or texture. As before, we consider the multimodal version of this problem, where the objective is to predict \emph{all} plausible images who scene layouts agree with the input segmentation map. This is a challenging problem, because only one image per scene layout is available -- in-the-wild collections of natural images usually do not contain multiple images with identical scene layouts.

\subsection{Relative Difficulty}

As previously noted in Section~\ref{sec:related_word_arb}, it is more challenging to generate multiple modes on the task of super-resolution than on the task of image synthesis from scene layout. The reason is that the factors of variation in the plausible output images largely do not depend on the input in the case of image synthesis from scene layout, whereas the factors of variation in the plausible output images depend very much on the input in the case of super-resolution. More concretely, in the case of image synthesis from scene layout, regardless of the input segmentation map, the output image could vary in terms of the time of the day and weather condition. On the other hand, in the case of super-resolution, the way the high-resolution output image varies depends on the regions in the low-resolution input image that are ambiguous. 

On the other hand, the different modes in the case of super-resolution are more similar than the different modes in the case of image synthesis from scene layout. This has two consequences: (1) it is easier to generate reasonable results for super-resolution without accounting for multimodality (i.e.: using a deterministic model, like a neural net without random noise input, and a regression loss), (2) it is more difficult to generate the different, subtle modes for super-resolution because explicitly encouraging diversity can easily result in spurious changes in the high-resolution image that are inconsistent with the low-resolution input image, such as global colour changes. The correct modes should instead be high-resolution images with subtle changes in the details that do not exist in the low-resolution input image; however, the magnitude of differences between these modes is actually smaller than that between spurious modes that exhibit global colour changes. 

\section{Experiments}
\label{seg:experiments}

\subsection{Single Image Super-Resolution}
\label{sec:srim_model}

\subsubsection{Architecture and Implementation Details}

\label{sec:srim_arch}
\begin{figure}
    \centering
    \includegraphics[width=0.8\textwidth]{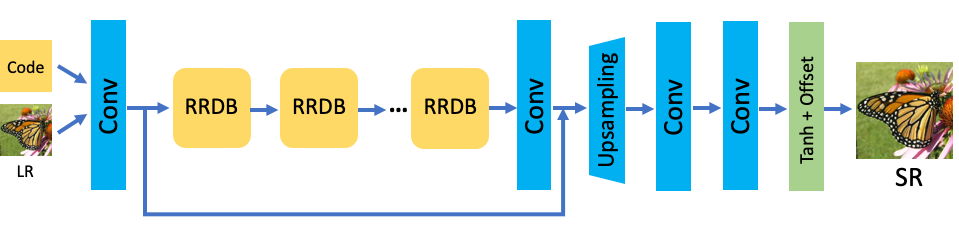}
    \caption{Network architecture for our super-resolution model.}
    \label{fig:srim_arch}
\end{figure}

The network architecture is based on residual-in-residual dense network in \citep{Wang2018ESRGANES}. We add an additional input for the latent vector $\mathbf{z}$, which is concatenated with the low-resolution input image before being fed into the first layer. In addition, we add a tanh activation, which produces an output in the range $[-1,1]$. We then scale the output of tanh and apply an offset to make the output lie in the range $[0, 1]$. The network architecture is shown in Figure~\ref{fig:srim_arch}. 

We first pretrain the network with the latent noise input $\mathbf{z}$ being set to zero. This is equivalent to turning the probabilistic model into a deterministic model, which usually trains faster. This comes at the cost of not modelling the inherent nondeterminism of the task (there is no single high-resolution image that corresponds to a low-resolution image), and so the best that this model can do is to predict the mean of all plausible high-resolution images, which generally looks reasonable but may be somewhat blurry (a problem that is sometimes known as ``regression to the mean''). To get around this issue, we revert to feeding random Gaussian latent noise to the model later on in training, which essentially turns the model back into a probabilistic model. Because the model is probabilistic now, it can learn the variations in the high-resolution image that are all plausible and will no longer predict the mean. In addition, we use nearest neighbour interpolation for the upsampling layer early on in training, which tends to produce images that are sharper but have more artifacts, and switch to bilinear upsampling later on. 

For the distance metric $\mathcal{L}(\cdot,\cdot)$, we use a linear combination of the $\ell_2$ distance between raw pixel values and the $\ell_2$ distance between the activations in the ``relu5\_4'' layer of the VGG-19 architecture. 

\subsubsection{Data}

For the problem of super-resolution, we can generate the training data by taking high-resolution images and downsampling them. The downsampled images will serve as input, and the original images will serve as the ground truth. The data is a subset of the ImageNet ILSVRC-2012 dataset. The ground truth images have a resolution of $256 \times 256$, which are obtained by anisotropic scaling of the original images. The input images have a resolution of $32 \times 32$, which are obtained by downsampling the $256 \times 256$ images. The images used for training and testing are disjoint. 

\begin{figure}
    \centering
    \begin{subfigure}{0.22\textwidth}
    \includegraphics[width=\textwidth]{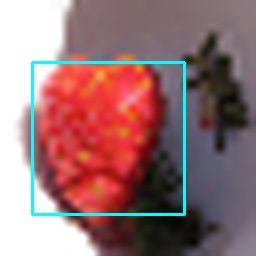}
    \caption{Input}
    \end{subfigure}
    \begin{subfigure}{0.33\textwidth}
    \includegraphics[width=\textwidth]{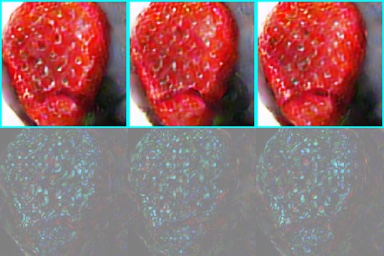}
    \caption{SRIM}
    \end{subfigure}
    \begin{subfigure}{0.33\textwidth}
    \includegraphics[width=\textwidth]{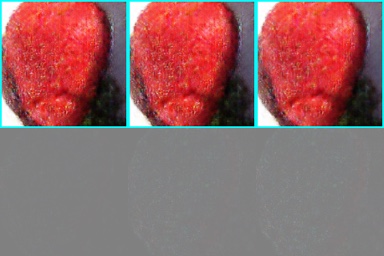}
    \caption{BicycleGAN}
    \end{subfigure}
    
    \centering
    \begin{subfigure}{0.22\textwidth}
    \includegraphics[width=\textwidth]{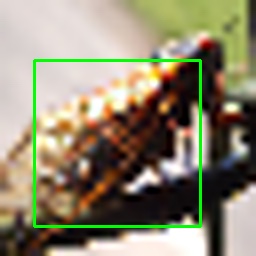}
    \caption{Input}
    \end{subfigure}
    \begin{subfigure}{0.33\textwidth}
    \includegraphics[width=\textwidth]{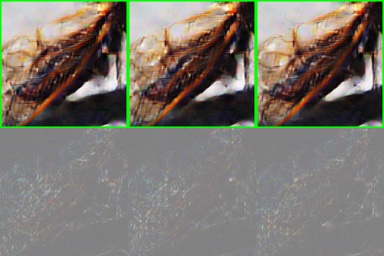}
    \caption{SRIM}
    \end{subfigure}
    \begin{subfigure}{0.33\textwidth}
    \includegraphics[width=\textwidth]{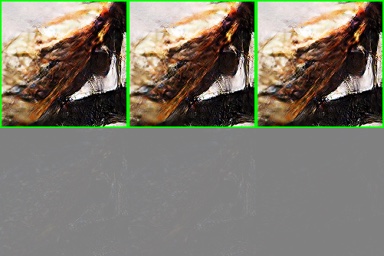}
    \caption{BicycleGAN}
    \end{subfigure}

    \centering
    \begin{subfigure}{0.22\textwidth}
    \includegraphics[width=\textwidth]{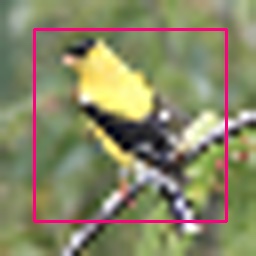}
    \caption{Input}
    \end{subfigure}
    \begin{subfigure}{0.33\textwidth}
    \includegraphics[width=\textwidth]{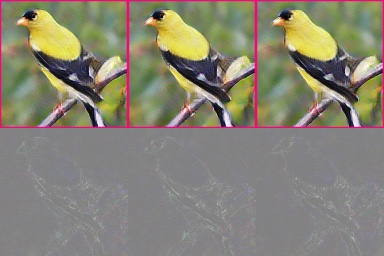}
    \caption{SRIM}
    \end{subfigure}
    \begin{subfigure}{0.33\textwidth}
    \includegraphics[width=\textwidth]{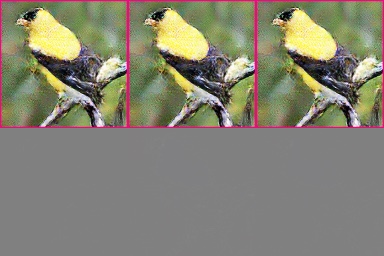}
    \caption{BicycleGAN}
    \end{subfigure}
    
    \caption{\label{fig:mult_comp} Samples generated by the proposed method (SRIM) and the baseline (BicycleGAN). The top row in each group of images shows different samples generated by each method, and the bottom row shows the difference between adjacent samples. As shown in the bottom row, the difference between the samples of SRIM is greater than that of BicycleGAN, which indicates that SRIM is able to generate more diverse samples. See \url{https://people.eecs.berkeley.edu/~ke.li/papers/imle_img_synth/fig4.gif} for a visualization of different samples.}
\end{figure}

\begin{figure}
    \centering
    \begin{subfigure}{0.22\textwidth}
    \includegraphics[width=\textwidth]{fig/sr_strawberry_input.jpg}
    \caption{Input}
    \end{subfigure}
    \begin{subfigure}{0.33\textwidth}
    \includegraphics[width=\textwidth]{fig/sr_strawberry_ours.jpg}
    \caption{SRIM}
    \end{subfigure}
    \begin{subfigure}{0.33\textwidth}
    \includegraphics[width=\textwidth]{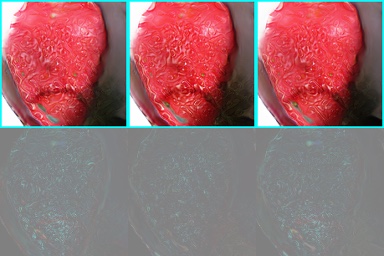}
    \caption{BicycleGAN with RRDB architecture}
    \end{subfigure}
    
    \centering
    \begin{subfigure}{0.22\textwidth}
    \includegraphics[width=\textwidth]{fig/sr_cicada_input.jpg}
    \caption{Input}
    \end{subfigure}
    \begin{subfigure}{0.33\textwidth}
    \includegraphics[width=\textwidth]{fig/sr_cicada_ours.jpg}
    \caption{SRIM}
    \end{subfigure}
    \begin{subfigure}{0.33\textwidth}
    \includegraphics[width=\textwidth]{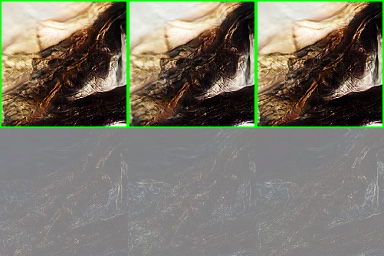}
    \caption{BicycleGAN with RRDB architecture}
    \end{subfigure}

    \centering
    \begin{subfigure}{0.22\textwidth}
    \includegraphics[width=\textwidth]{fig/sr_bird_input.jpg}
    \caption{Input}
    \end{subfigure}
    \begin{subfigure}{0.33\textwidth}
    \includegraphics[width=\textwidth]{fig/sr_bird_ours.jpg}
    \caption{SRIM}
    \end{subfigure}
    \begin{subfigure}{0.33\textwidth}
    \includegraphics[width=\textwidth]{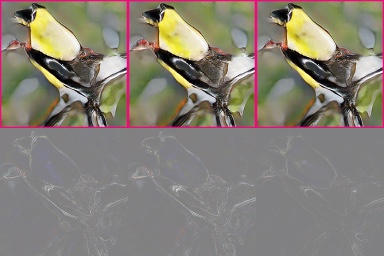}
    \caption{BicycleGAN with RRDB architecture}
    \end{subfigure}
    
    \caption{\label{fig:mult_bgan} Samples generated by the proposed method (SRIM) and the baseline (BicycleGAN with RRDB), which shares the same generator network architecture as the proposed method. The top row in each group of images shows different samples generated by each method, and the bottom row shows the difference between adjacent samples. As shown, the results of BicycleGAN with RRDB are not consistent with the input. Specifically, the colours are not consistent, e.g.: the surface of the strawberry, the body of the insect, and the background to the right of the bird. On the other hand, SRIM does not have colour hallucinations and is faithful to the input.
    }
\end{figure}

\begin{figure}
    \centering
    \begin{subfigure}{0.22\textwidth}
    \includegraphics[width=\textwidth]{fig/sr_strawberry_input.jpg}
    \caption{Input}
    \end{subfigure}
    \begin{subfigure}{0.33\textwidth}
    \includegraphics[width=\textwidth]{fig/sr_strawberry_ours.jpg}
    \caption{SRIM}
    \end{subfigure}
    \begin{subfigure}{0.33\textwidth}
    \includegraphics[width=\textwidth]{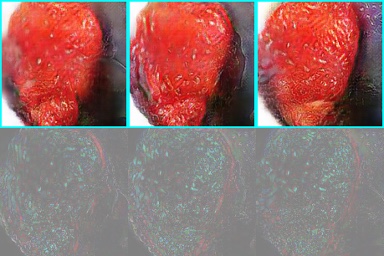}
    \caption{MR-GAN}
    \end{subfigure}
    
    \centering
    \begin{subfigure}{0.22\textwidth}
    \includegraphics[width=\textwidth]{fig/sr_cicada_input.jpg}
    \caption{Input}
    \end{subfigure}
    \begin{subfigure}{0.33\textwidth}
    \includegraphics[width=\textwidth]{fig/sr_cicada_ours.jpg}
    \caption{SRIM}
    \end{subfigure}
    \begin{subfigure}{0.33\textwidth}
    \includegraphics[width=\textwidth]{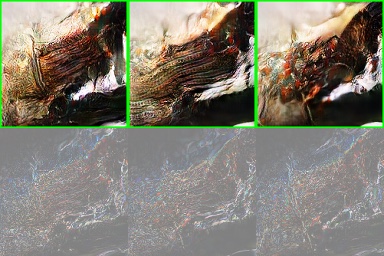}
    \caption{MR-GAN}
    \end{subfigure}

    \centering
    \begin{subfigure}{0.22\textwidth}
    \includegraphics[width=\textwidth]{fig/sr_bird_input.jpg}
    \caption{Input}
    \end{subfigure}
    \begin{subfigure}{0.33\textwidth}
    \includegraphics[width=\textwidth]{fig/sr_bird_ours.jpg}
    \caption{SRIM}
    \end{subfigure}
    \begin{subfigure}{0.33\textwidth}
    \includegraphics[width=\textwidth]{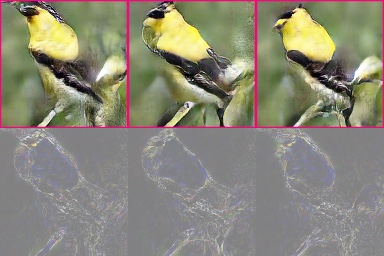}
    \caption{MR-GAN}
    \end{subfigure}
    
    \caption{\label{fig:mult_mrgan} Samples generated by the proposed method (SRIM) and the baseline (MR-GAN). The top row in each group of images shows different samples generated by each method, and the bottom row shows the difference between adjacent samples. As shown, in addition to being inconsistent with the input, MR-GAN results contain obvious artifacts, e.g.: on the left portion of the strawberry, on the wing of the bug and on the tail of the bird. SRIM has far fewer artifacts and is faithful to the input.
    }
\end{figure}

\subsubsection{Baselines}

Because there has been no prior work on multimodal super-resolution, we compare to leading generic multimodal image synthesis methods, BicycleGAN~\citep{DBLP:journals/corr/abs-1711-11586}
and MR-GAN~\citep{Lee2019HarmonizingML}. We use the official implementation for all baselines. For BicycleGAN, we use two different generator architectures, the architecture used in the official implementation, and the architecture used for the proposed method, which we denote as BicycleGAN with RRDB. 

\subsubsection{Experimental Setting}
We train our model on 1280 training images. We use 5 noise channels and set the hyperparameters shown in Algorithm \ref{alg:cimle} to the following values: $|S|=640$, $m=100$, $L=200000$, $|\widetilde{S}|=4$ and $\eta=10^{-5}$. 

\subsubsection{Evaluation Metric}

We would like a multimodal method to produce samples that are both diverse and consistent with the low-resolution input image. We propose a evaluation metric, which we call faithfulness-weighted variance, that captures both of these properties. It is defined as follows:

\[
\mathcal{M} = \sum_i\sum_j w_i d_{\mathrm{LPIPS}}(\widetilde{\mathbf{y}}_{i,j}, \bar{\mathbf{y}}_{j}), \mbox{where }
w_i = \exp\left(-\frac{d_{\mathrm{LPIPS}}(\widetilde{\mathbf{y}}_{i,j}, {\mathbf{y}}_{j})}{2\sigma^2}\right)
\]
Here, $\widetilde{\mathbf{y}}_{i,j}$ denotes the $i^{th}$ generated sample conditioned on the $j^{th}$ input image, $\bar{\mathbf{y}}_{j}$ denotes the mean sample conditioned on the $j^{th}$ input image, $\mathbf{y}_{j}$ denotes the ground truth high-resolution image corresponding to the $j^{th}$ input image, and $d_{\mathrm{LPIPS}}(\cdot,\cdot)$ denotes the LPIPS distance metric~\citep{zhang2018unreasonable}. The $w_i$ is the faithfulness coefficient which is obtained by passing the LPIPS distance between the generated sample and the ground truth image through a Gaussian kernel. We report results on a range of different bandwidth parameters for the Gaussian kernel. Intuitively, the faithfulness coefficient ensures that a sample only contributes significantly to the metric if it is perceptually consistent with the ground truth image. 

\subsubsection{Findings}

\begin{table}
  \centering
  \begin{tabular}{lcccc}
    \toprule
    {$\sigma$} & {BicycleGAN} & {BicycleGAN with RRDB} & {MR-GAN} & {SRIM}\\
      \midrule
    0.3 & 4.31$ \times 10^{-2}$ & 3.00$ \times 10^{-2}$ & 4.82$ \times 10^{-2}$ & $\mathbf{5.91 \times 10^{-2}}$ \\
    0.2 & 5.19$ \times 10^{-3}$ & 2.17$ \times 10^{-3}$ & 2.62$ \times 10^{-3}$ & $\mathbf{9.23 \times 10^{-3}}$ \\
    0.15 & 4.67$ \times 10^{-4}$ & 1.14$ \times 10^{-4}$ & 8.90$ \times 10^{-5}$ & $\mathbf{1.02 \times 10^{-3}}$ \\
    \bottomrule
  \end{tabular}
  \caption{Comparison of faithfulness weighted variance achieved by BicycleGAN, BicycleGAN with the same generator architecture (RRDB), MR-GAN and the proposed method (SRIM) . Higher value indicates richer variation in the generated samples.}
\label{tab:weighted_var}
\end{table}

We applied the metric to both the proposed method (SRIM) and the baselines to a test set with 80 unseen images and 50 samples are drawn for each input. The average faithfulness-weighted variance are reported in Table~\ref{tab:weighted_var}. As shown, SRIM achieved a higher faithfulness-weighted variance compared to the baselines, which indicates it is capable of generating more diverse results while being also faithful to the target image. This is reflected in the qualitative results in Figures~\ref{fig:mult_comp}, \ref{fig:mult_bgan} and \ref{fig:mult_mrgan}, which demonstrate greater differences between different samples compared to the baselines. 

\subsection{Image Synthesis from Semantic Layout}
\label{sec:image_synthesis}

\subsubsection{Architecture and Implementation Details}\label{sec:crn}

The network architecture is based on the Cascaded Refinement Network (CRN) architecture~\citep{chen2017photographic}. The vanilla CRN synthesizes only one image for the same semantic layout input. To endow the network with the capability to generate an arbitrary number of modes, we add additional input channels to the architecture and feed random noise $\mathbf{z}$ via these channels. Because the noise is random, the neural net can now be viewed as a (implicit) probabilistic model, which we can train using conditional IMLE. 

Because the input segmentation maps are provided at high resolutions, the noise vector $\mathbf{z}$, which is concatenated to the input channel-wise, could be very high-dimensional, which could hurt sample efficiency and therefore training speed. To solve this, we propose forcing the noise to lie on a low-dimensional manifold. To this end, we add a noise encoder module, which is a 3-layer convolutional net that takes the segmentation $\mathbf{x}$ and a lower-dimensional noise vector $\widetilde{\mathbf{z}}$ as input and outputs a noise vector $\mathbf{z}'$ of the same size as $\mathbf{z}$. We replace $\mathbf{z}$ with $\mathbf{z}'$ and leave the rest of the architecture unchanged.

For the distance metric $\mathcal{L}(\cdot,\cdot)$, we use a linear combination of $\ell_1$ distances between the activations in the following layers of the VGG-19 architecture: ``conv1\_2'', ``conv2\_2'', ``conv3\_2'', ``conv4\_2'' and ``conv5\_2''. 

\subsubsection{Data}
\begin{figure}
    \centering
    \includegraphics[width=0.45\linewidth]{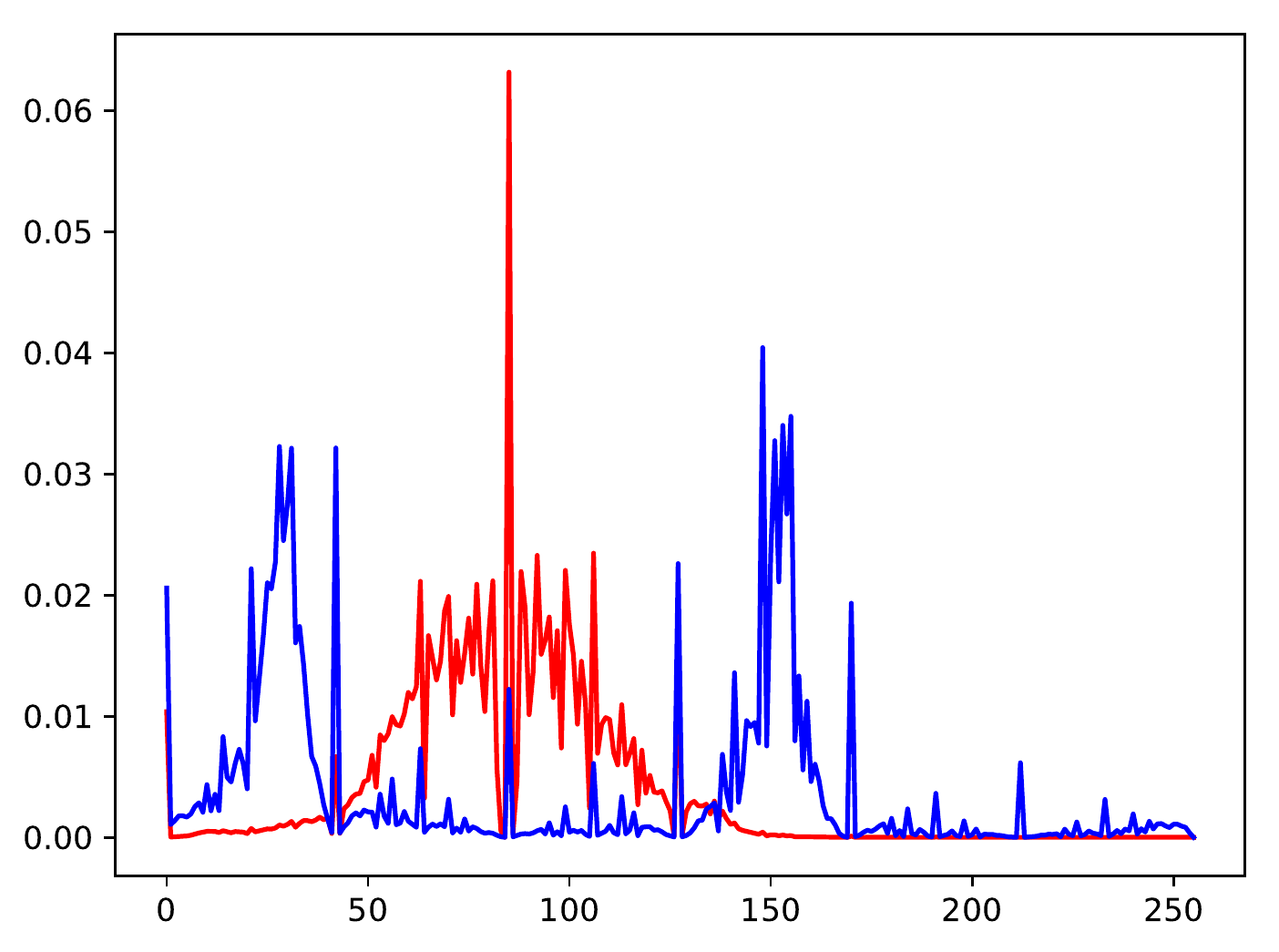}
    \caption{Comparison of histogram of hues between two datasets. Red is Cityscapes and blue is GTA-5. }
    \label{fig:hist}
\end{figure}

\begin{figure}[htbp]
    \centering
    \begin{subfigure}{0.4\linewidth}
    \includegraphics[width=\linewidth]{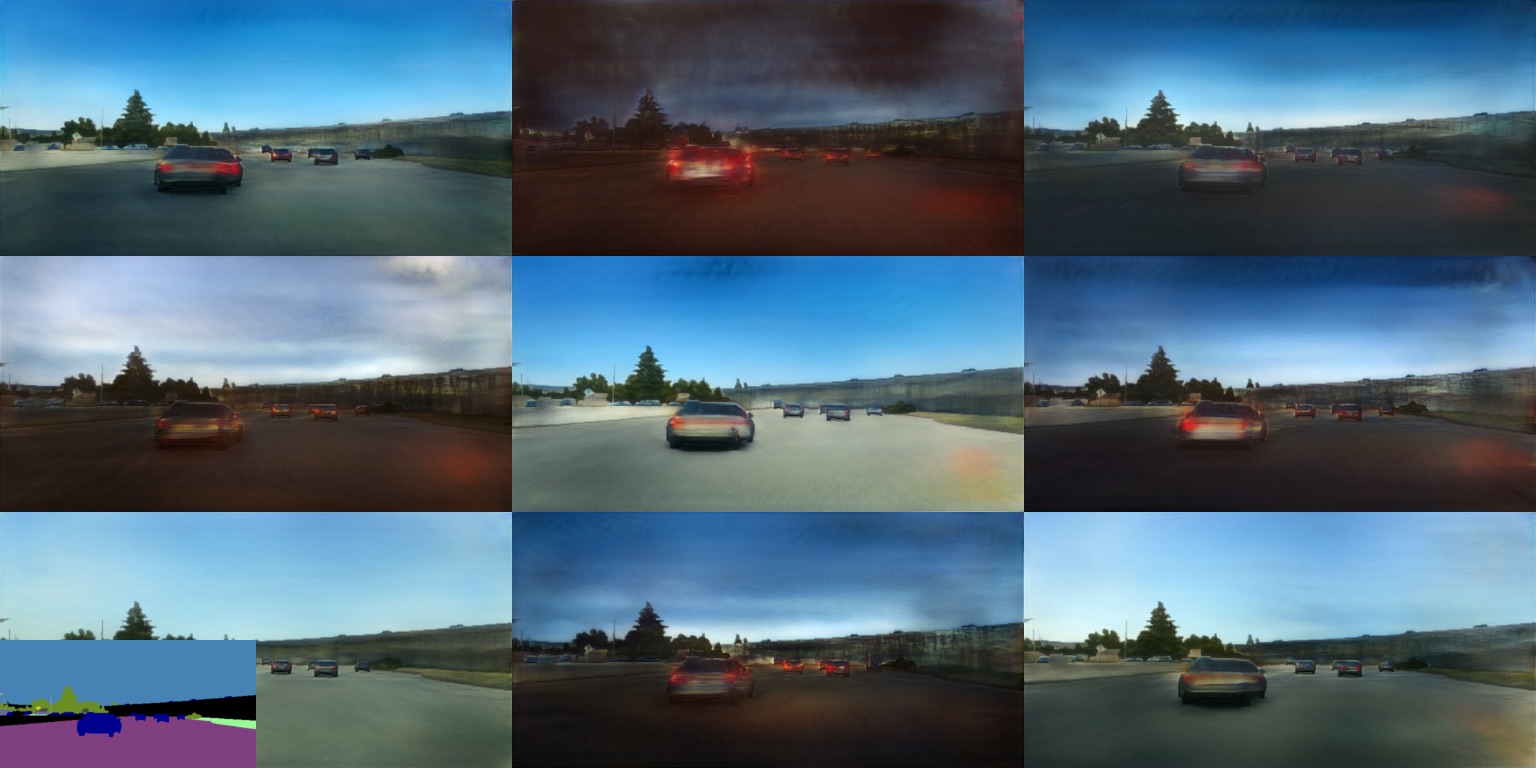}
    \caption{}
    \end{subfigure}
    \begin{subfigure}{0.4\linewidth}
    \includegraphics[width=\linewidth]{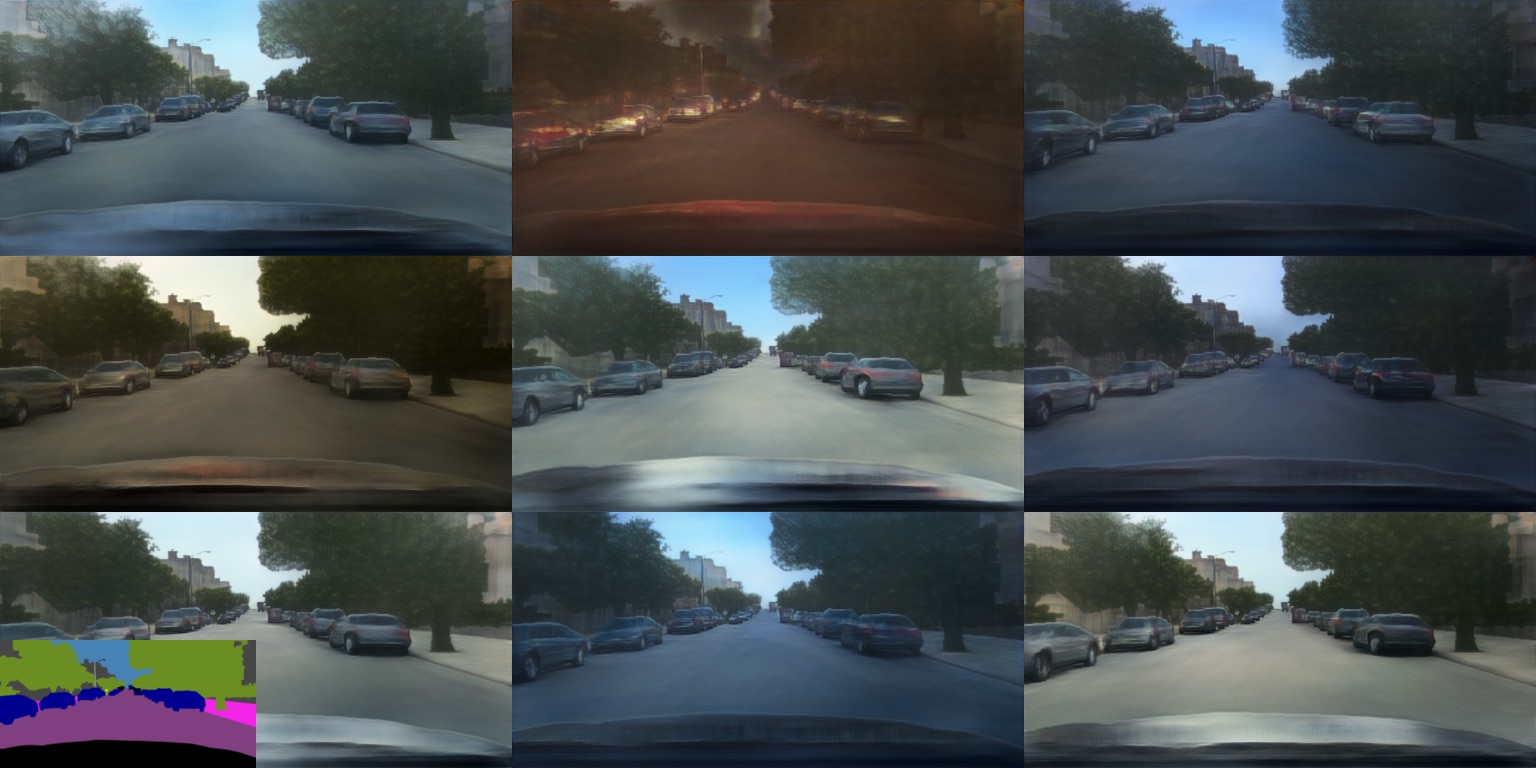}
    \caption{}
    \end{subfigure}
    \caption{Images generated using our method on the BDD100K dataset.}
    \label{fig:bdd}
\end{figure}

The choice of dataset is very important for multimodal conditional image synthesis -- if the training dataset does not contain much variation, then there is no hope of synthesizing images that do. The most common dataset in the unimodal setting is the Cityscapes dataset \citep{Cordts2016Cityscapes}. However, it is not suitable for the multimodal setting because most images in the dataset are taken under similar weather conditions and time of day and the amount of variation in object colours is limited. This lack of diversity limits what any multimodal method can do. On the other hand, the GTA-5 dataset~\citep{Richter_2016_ECCV}, has much greater variation in terms of weather conditions and object appearance. To demonstrate this, we compare the colour distribution of both datasets and present the distribution of hues of both datasets in Figure~\ref{fig:hist}. As shown, Cityscapes is concentrated around a single mode in terms of hue, whereas GTA-5 has much greater variation in hue. Additionally, the GTA-5 dataset includes more 20000 images and so is much larger than Cityscapes. As a result, we use GTA-5 as our main source of training data. To demonstrate the generalizability of our approach and its applicability to real-world datasets, we also train on the BDD100K~\citep{yu2018bdd100k} dataset and show results in Fig. \ref{fig:bdd}.

\subsubsection{Baselines}

We compare to two specialized methods that are designed for generating images from scene layouts, Pix2pix-HD~\citep{wang2017high} and CRN~\citep{chen2017photographic}. We consider variants of each that are adapted for the multimodal setting. In the case of Pix2pix-HD, we consider a generator that takes in both the segmentation map and noise as input (so that it could in principle produce different images given the same segmentation map), which we will refer to as Pix2pix-HD with input noise. In the case of CRN, we consider the variant proposed in \citep{chen2017photographic} with 27 output channels that can generate nine images from the same segmentation map. Additionally, as in the case of super-resolution, we compare to leading generic multimodal image synthesis methods like BicycleGAN~\citep{DBLP:journals/corr/abs-1711-11586} and MR-GAN~\citep{Lee2019HarmonizingML}. For BicycleGAN, we use two different generator architectures, the architecture used in the official implementation, and the architecture used for the proposed method (CRN), which we denote as BicycleGAN with CRN architecture. 

\subsubsection{Rebalancing}

Datasets are usually strongly biased towards objects with relatively common appearance, and so even if the model can learn the distribution of images that underlies the dataset, samples will not necessarily be diverse in terms of appearance. To learn a model that generates diverse samples, we need to reweight different images in the dataset and also different objects within images so that rare images and objects are upweighted. We propose two strategies to do so:

\paragraph{Dataset Rebalancing} We first rebalance the dataset to increase the chance of rare images being sampled when populating the training batch $S$. To this end, for each training image, we calculate the average colour of each category in that image. Then for each category, we estimate the distribution of the average colour of the category over different training images using a Gaussian kernel density estimate (KDE). 

More concretely, we compute the average colour of each semantic category $p$ for each image $k$, which is a three-dimensional vector:
\begin{equation*}
    \mathbf{c}_{k}(p)= \frac{\sum_{i=1}^{h}\sum_{j=1}^{w} \mathbf{1}\left[\mathbf{x}_k^{i,j}=p\right] \mathbf{y}_k^{i,j}}{\sum_{i=1}^{h}\sum_{j=1}^{w} \mathbf{1}\left[\mathbf{x}_k^{i,j}=p\right]}
\end{equation*}

For each category $p$, we consider the set of average colours for that category in all training images, i.e.: $\{\mathbf{c}_{k}(p) \vert k \in \{1,\ldots,n\}\text{ such that class }p\text{ appears in }\mathbf{x}_k\}$. We then fit a Gaussian kernel density estimate to this set of vectors and obtain an estimate of the distribution of average colours of category $p$. Let $D_p(\cdot)$ denote the estimated probability density function (PDF) for category $p$. We define the \textit{rarity score} of category $p$ in the $k^{\mathrm{th}}$ training image as follows:
\begin{equation*}
    R_p(k) = \begin{cases}
    \frac{1}{D_p(\mathbf{c}_{k}(p))} &\text{class } p \text{ appears in } \mathbf{x}_k\\
    0 &\text{otherwise}
    \end{cases}
\end{equation*}

When populating training batch $S$, we allocate a portion of the batch to each of the top five categories that have the largest overall area across the dataset. For each category, we sample training images with a probability in proportion of their rarity scores. Effectively, we upweight images containing objects with rare appearance. 

The rationale for selecting the categories with the largest areas is because they tend to appear more frequently and be visually more prominent. If we were to allocate a fixed portion of the batch to rare categories, we would risk overfitting to images containing those categories. 

\paragraph{Loss Rebalancing} The same training image can contain both common and rare objects. Therefore, we modify the loss function so that the objects with rare appearance are upweighted. For each training pair $(\mathbf{x}_k,\mathbf{y}_k)$, we define a rarity score mask $\mathcal{M}^k\in \mathbb{R}^{h\times w\times 1}$:
\begin{equation*}
    \mathcal{M}^k_{i,j} = R_p(k)\quad \text{if pixel }(i,j) \text{ belongs to class }p
\end{equation*}
We then normalize $\mathcal{M}^k$ so that every entry lies in $(0,1]$:
\begin{equation*}
    \widehat{\mathcal{M}}^k = \frac{1}{\max_{i,j} \mathcal{M}^k_{i,j}}\mathcal{M}^k
\end{equation*}
The mask is then used to weight different pixels differently the loss function~\eqref{eq:loss}. Let $\widehat{\mathcal{M}}$ be the normalized rarity score mask associated with the training pair $(\mathbf{x},\mathbf{y})$. The new loss $\mathcal{L}$ becomes:
\begin{equation*}
    \mathcal{L}(\mathbf{y}, \widetilde{\mathbf{y}}) = \sum_{i=1}^l\lambda_l\left\Vert\widehat{\mathcal{M}}_i\circ\left[\Phi_i(\mathbf{y})-\Phi_i(\widetilde{\mathbf{y}})\right]\right\Vert_1
\end{equation*}
Here $\widehat{\mathcal{M}}_i$ is the normalized rarity score mask $\widehat{\mathcal{M}}$ downsampled to match the size of $\Phi_i(\cdot)$, and $\circ$ denotes the element-wise product. 

\begin{figure*}
    \centering
    \begin{subfigure}{0.47\textwidth}
    \includegraphics[width=\linewidth]{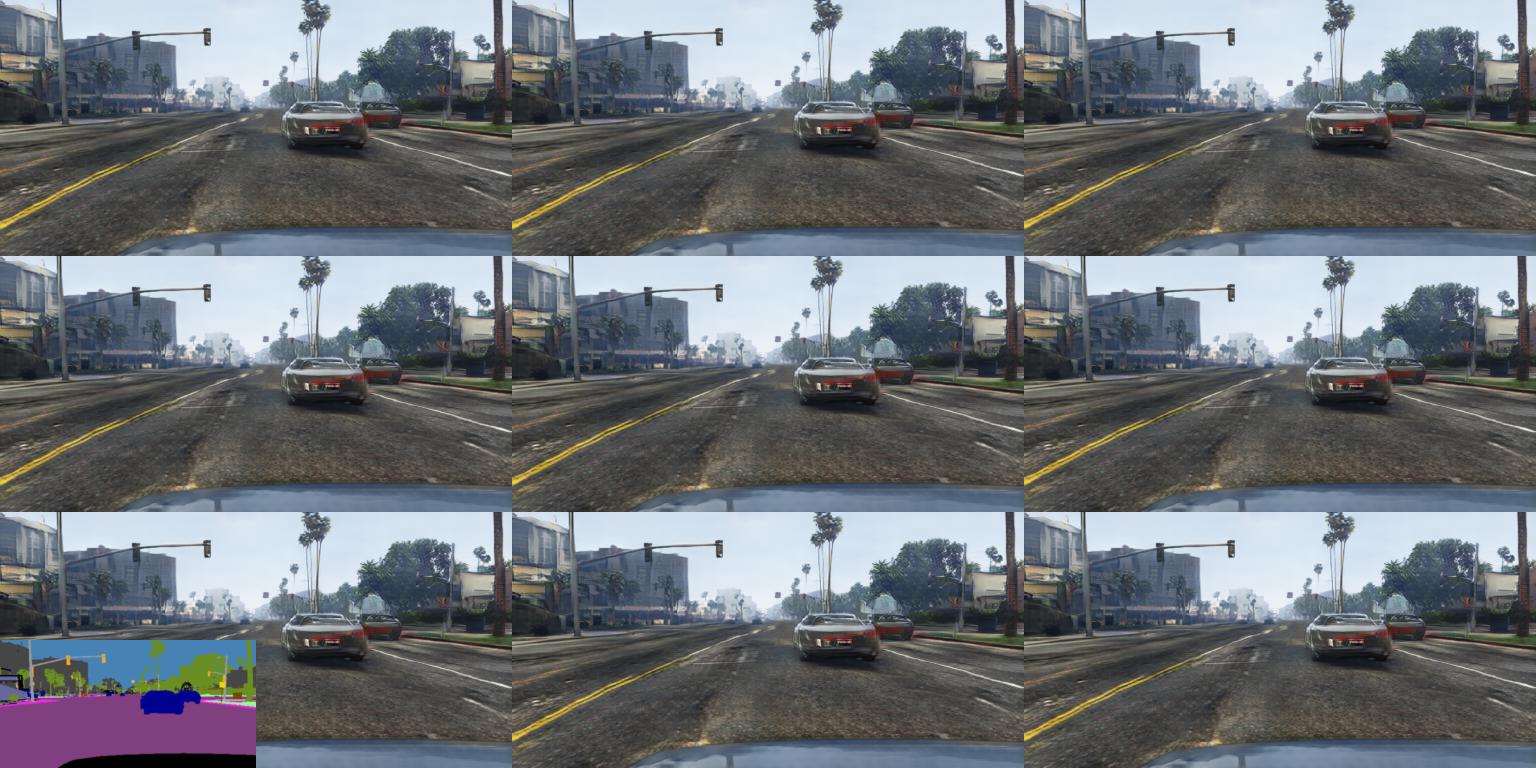}
    \caption{Pix2pix-HD+noise}
    \end{subfigure}
    \begin{subfigure}{0.47\textwidth}
    \includegraphics[width=\linewidth]{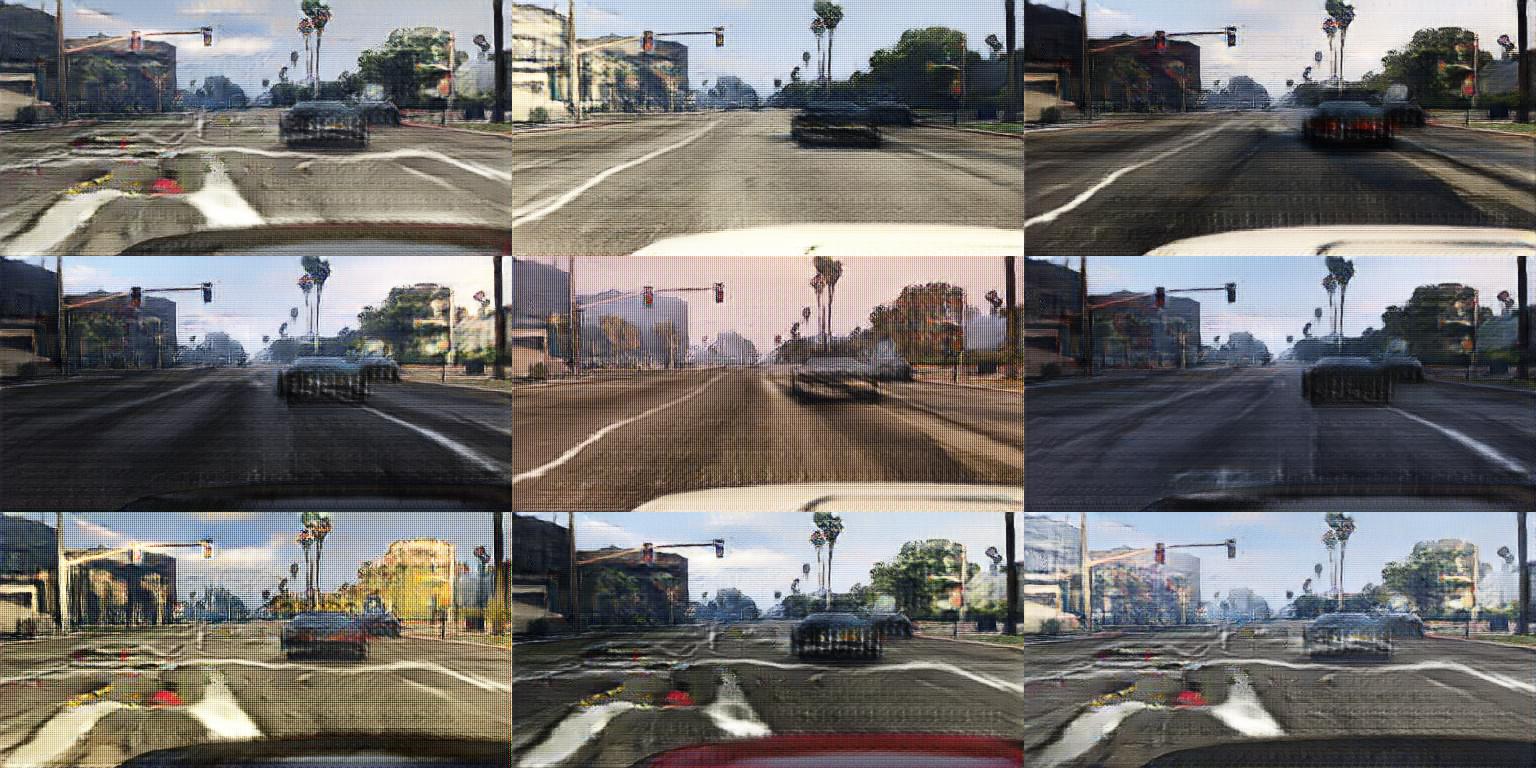}
    \caption{BicycleGAN}
    \end{subfigure}
    \begin{subfigure}{0.47\textwidth}
    \includegraphics[width=\linewidth]{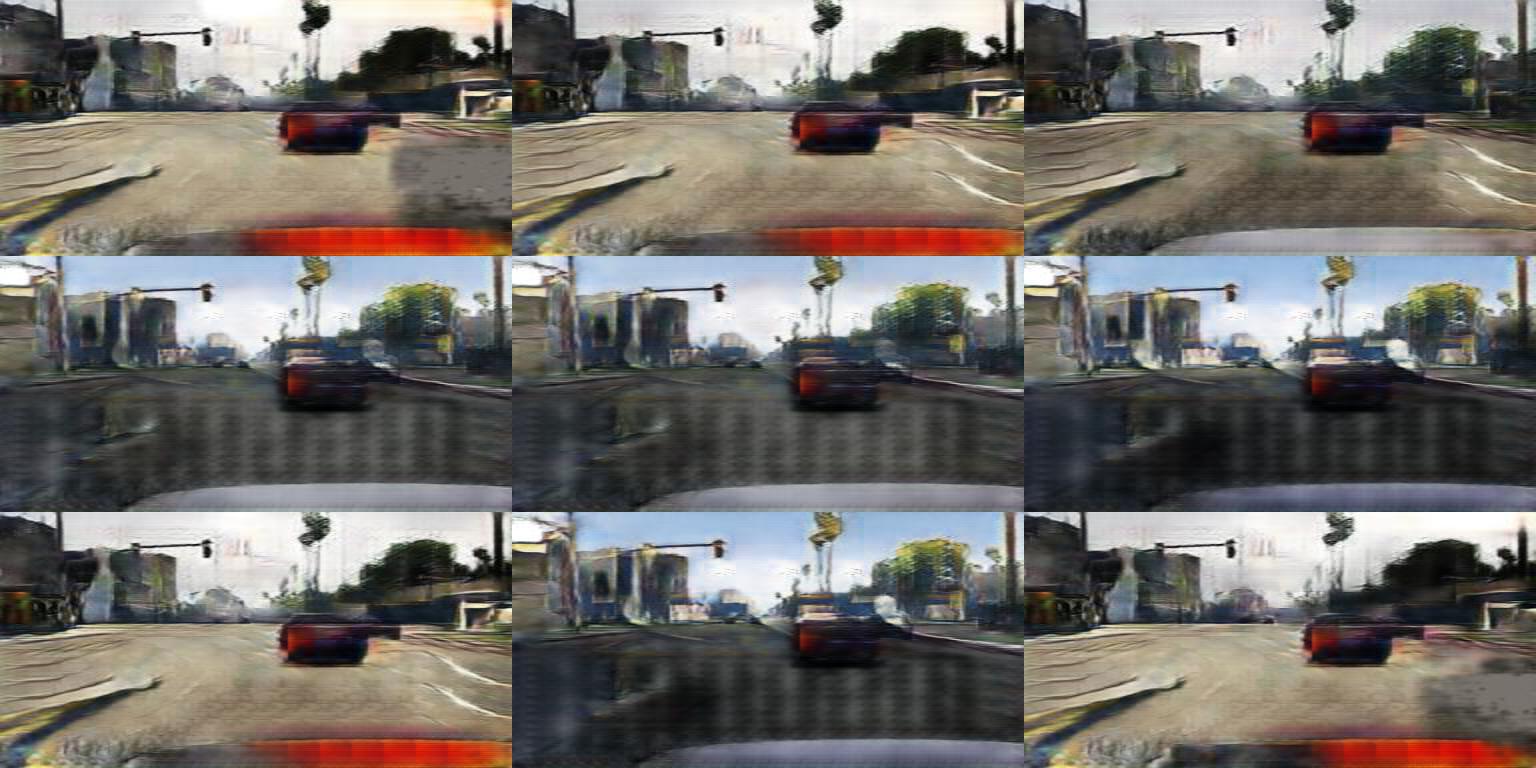}
    \caption{MR-GAN}
    \end{subfigure}
    \begin{subfigure}{0.47\textwidth}
    \includegraphics[width=\linewidth]{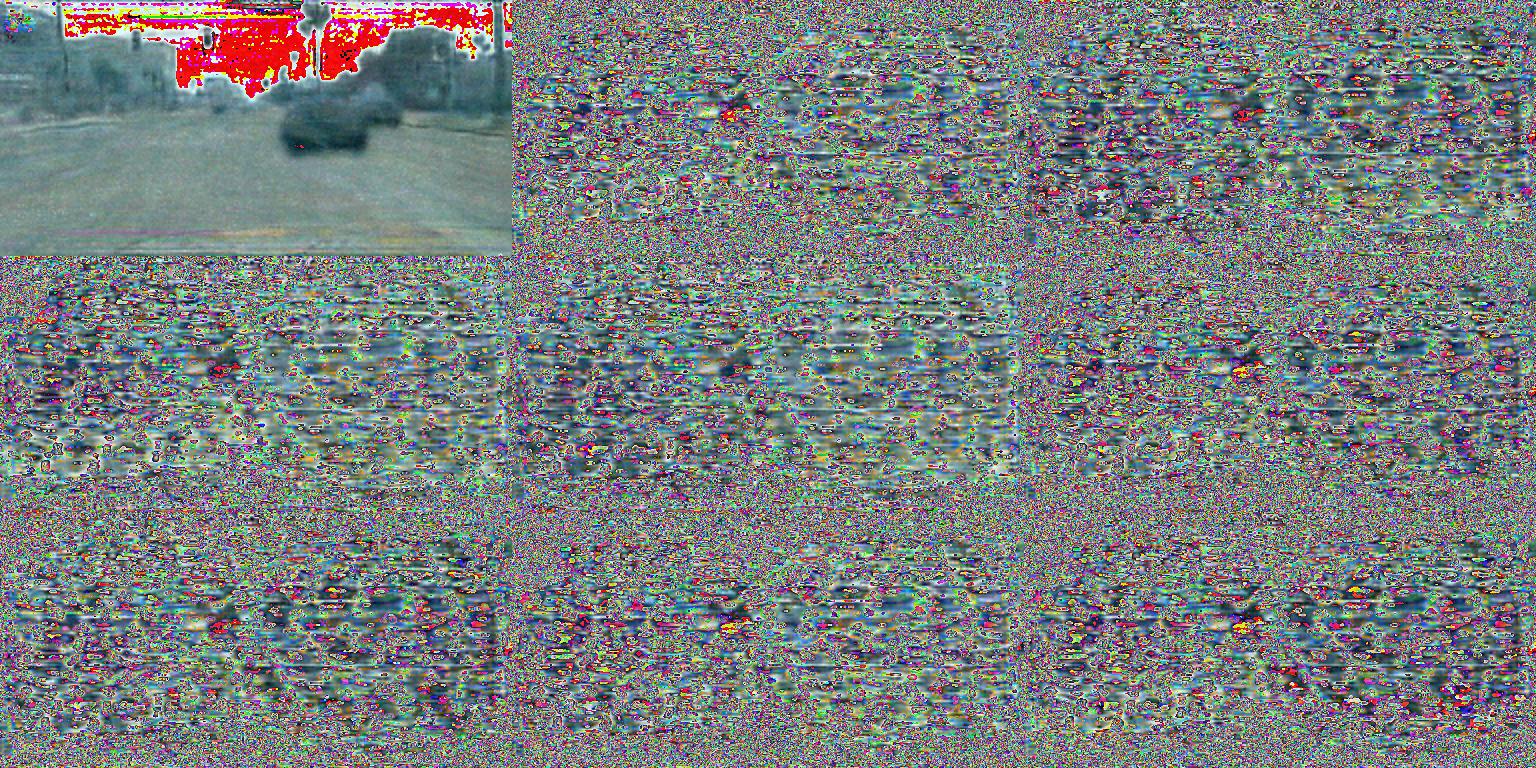}
    \caption{BicycleGAN with CRN architecture}
    \end{subfigure}
    \begin{subfigure}{0.47\textwidth}
    \includegraphics[width=\linewidth]{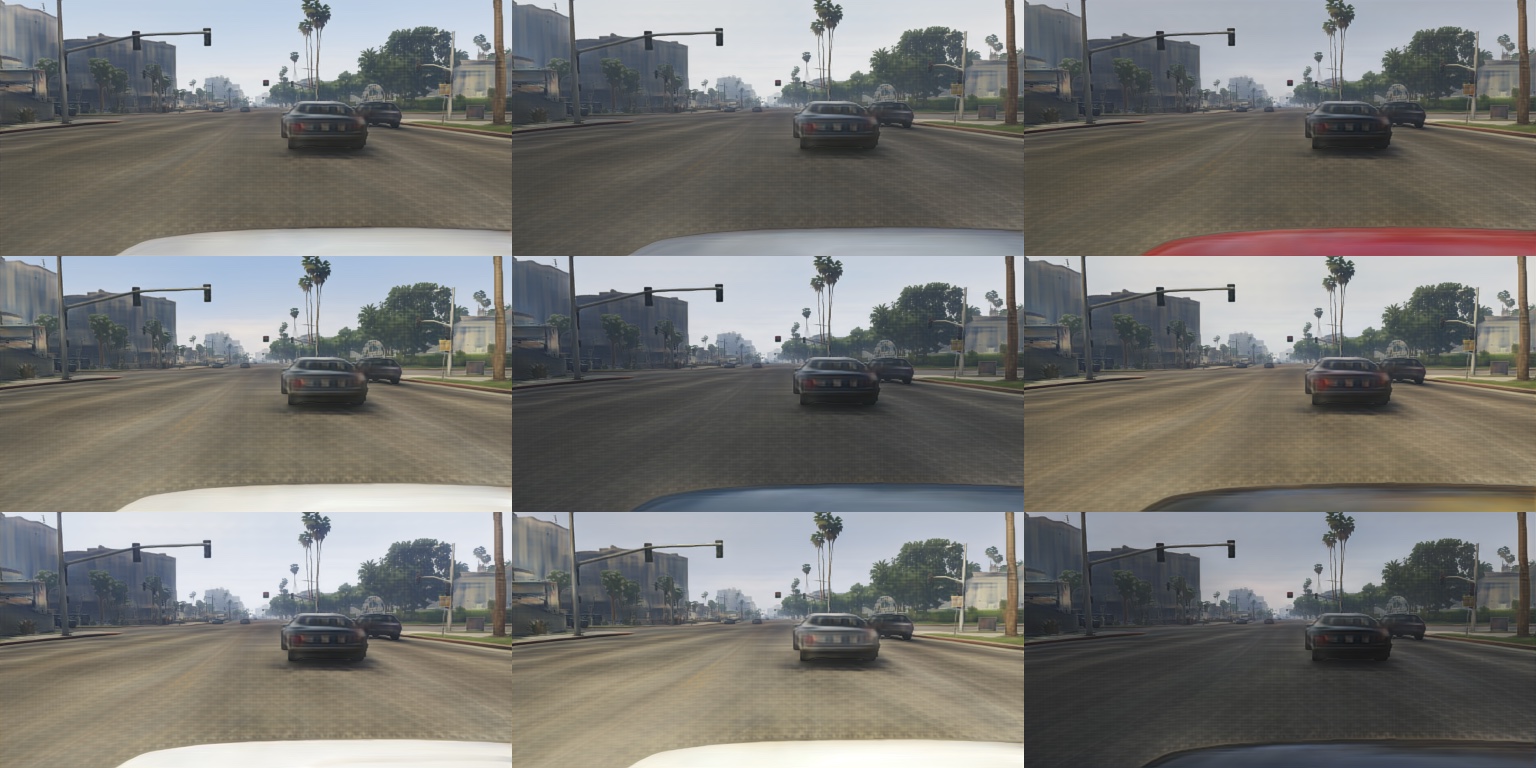}
    \caption{CRN}
    \end{subfigure}
    \begin{subfigure}{0.47\textwidth}
    \includegraphics[width=\linewidth]{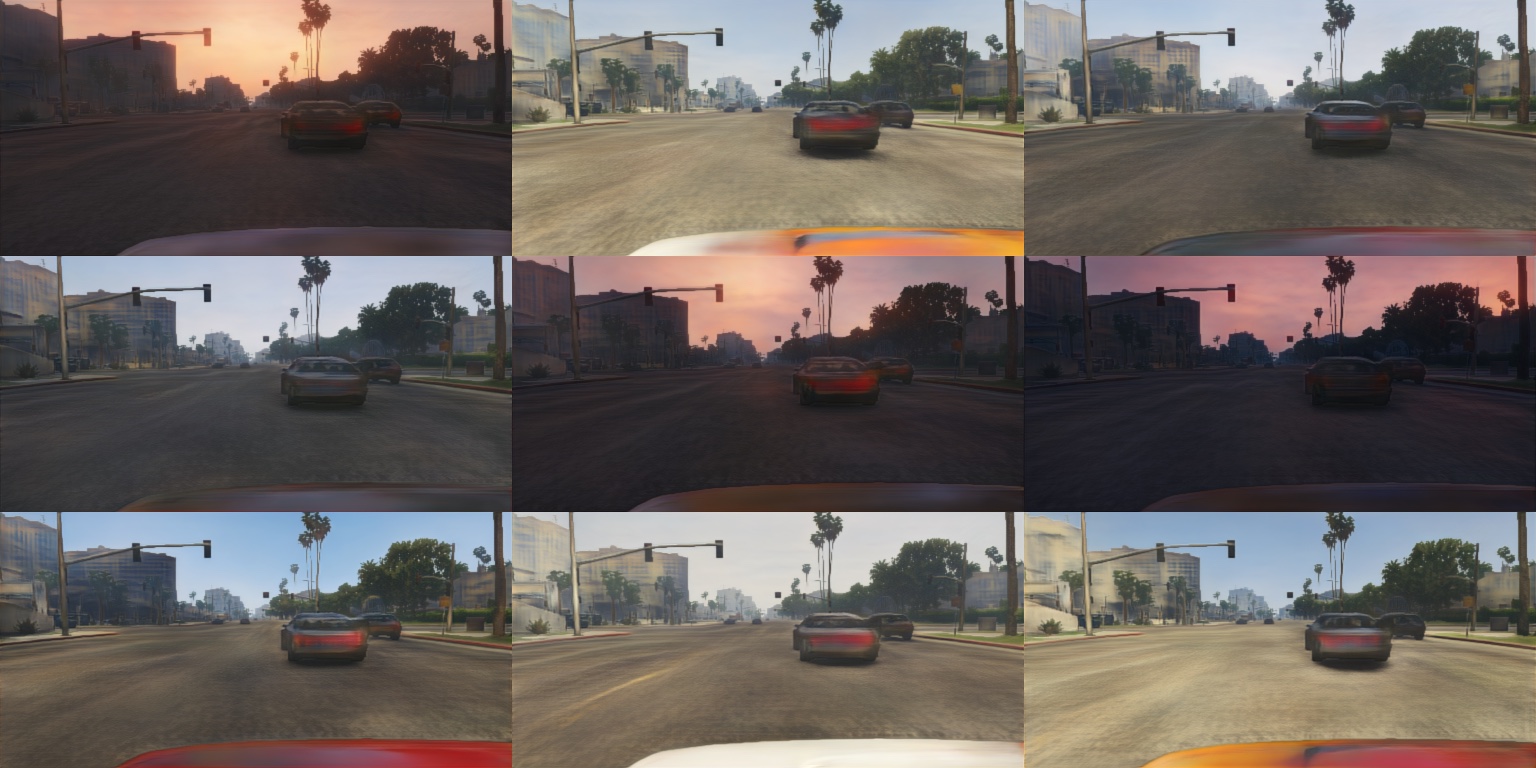}
    \caption{Our model}
    \end{subfigure}
    \caption{Comparison of generated images for the same semantic layout. The bottom-left image in (a) is the input semantic layout and we generate 9 samples for each model. See \url{https://people.eecs.berkeley.edu/~ke.li/papers/imle_img_synth/fig6.gif} for a visualization of different samples. Note that BicycleGAN training becomes unstable when the generator architecture is replaced with CRN, and despite weeks of hyperparameter tuning, we could not get reasonable results. The difficulty of getting BicycleGAN to work with CRN architecture is confirmed by the authors of BicycleGAN. }
    \label{fig:samp}
\end{figure*}
\begin{figure*}
    \centering
    \begin{subfigure}{0.47\textwidth}
    \includegraphics[width=\linewidth]{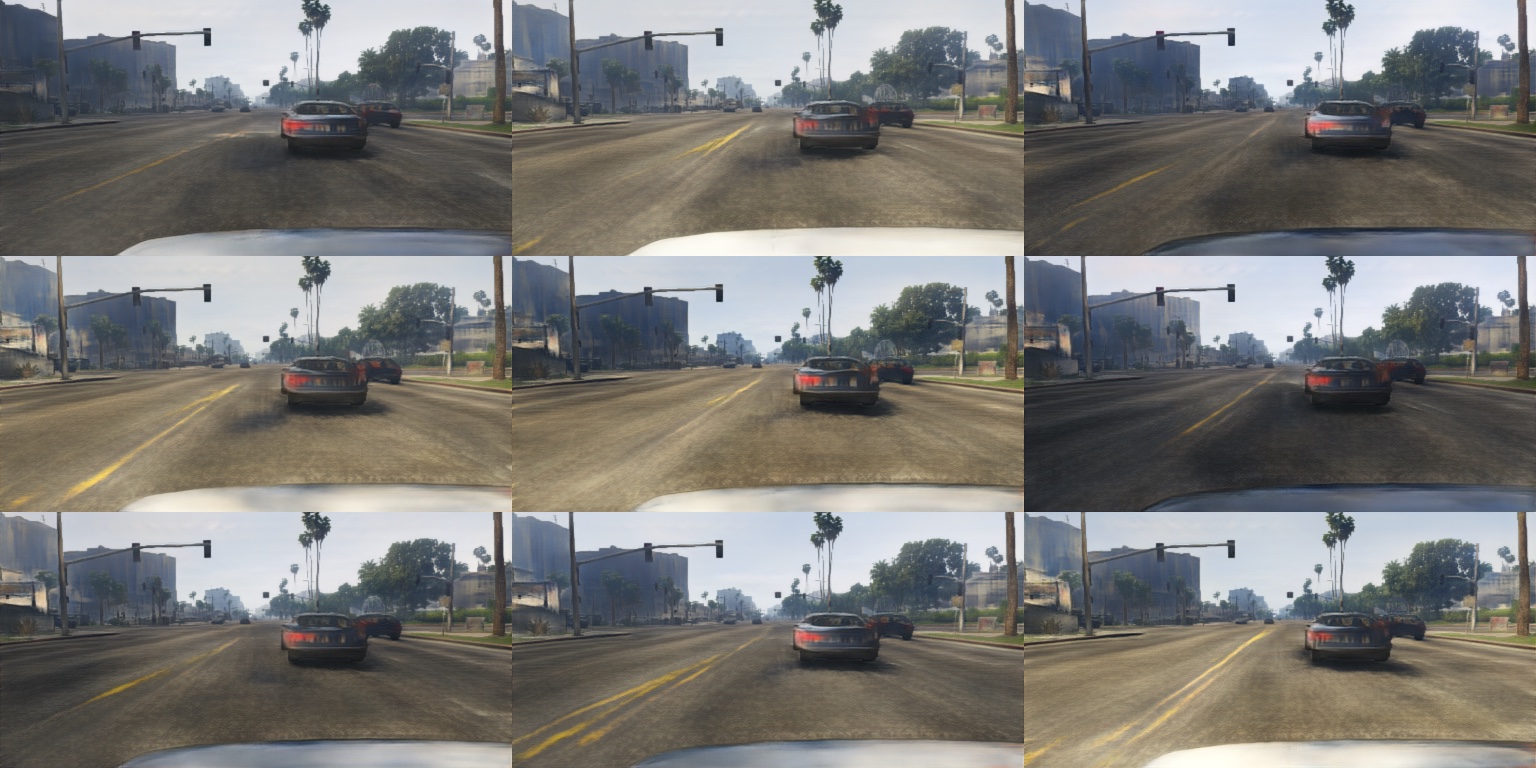}
    \caption{Our model w/o the noise encoder and rebalancing scheme}
    \end{subfigure}
    \begin{subfigure}{0.47\textwidth}
    \includegraphics[width=\linewidth]{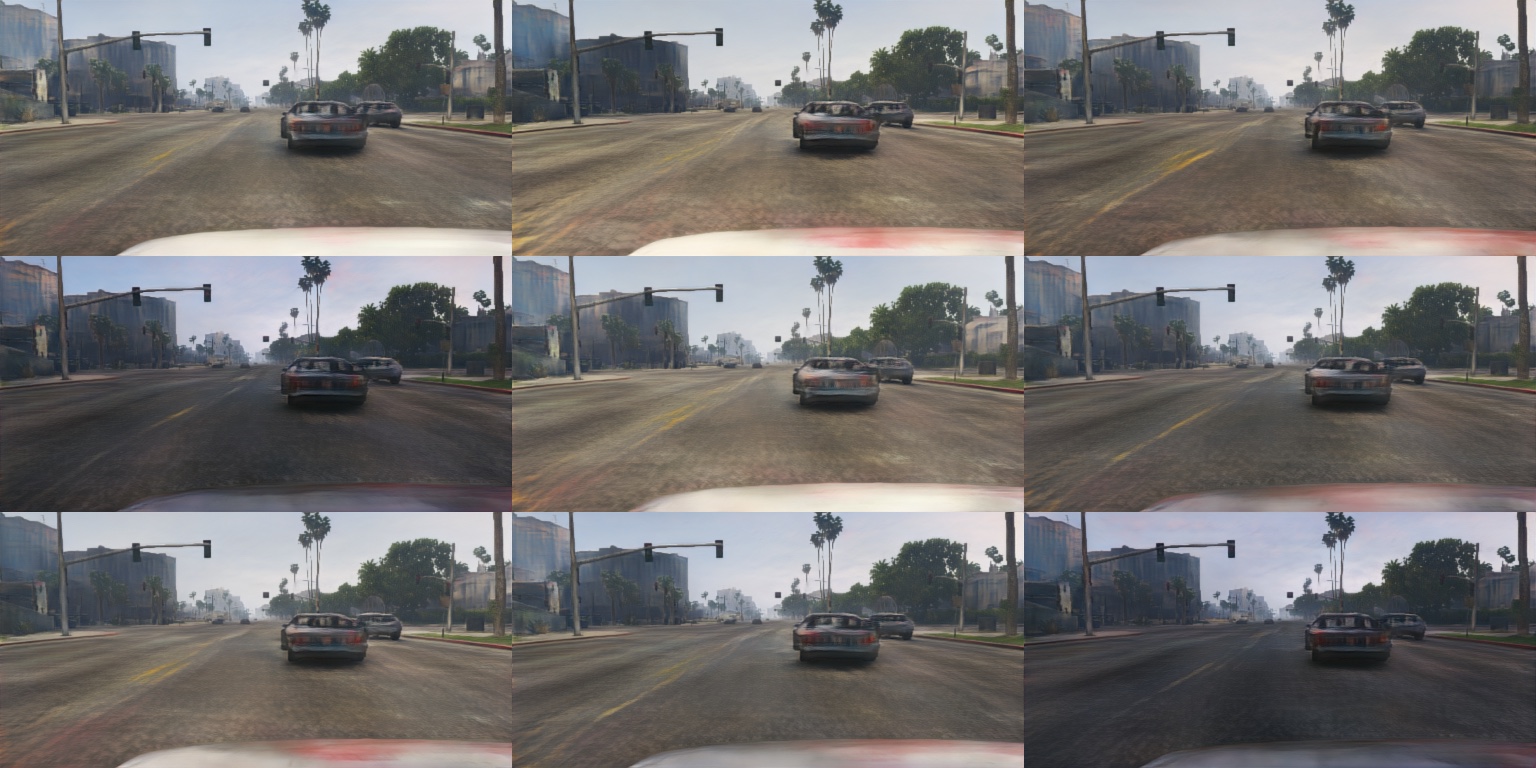}
    \caption{Our model w/o the noise encoder}
    \end{subfigure}
    \begin{subfigure}{0.47\textwidth}
    \includegraphics[width=\linewidth]{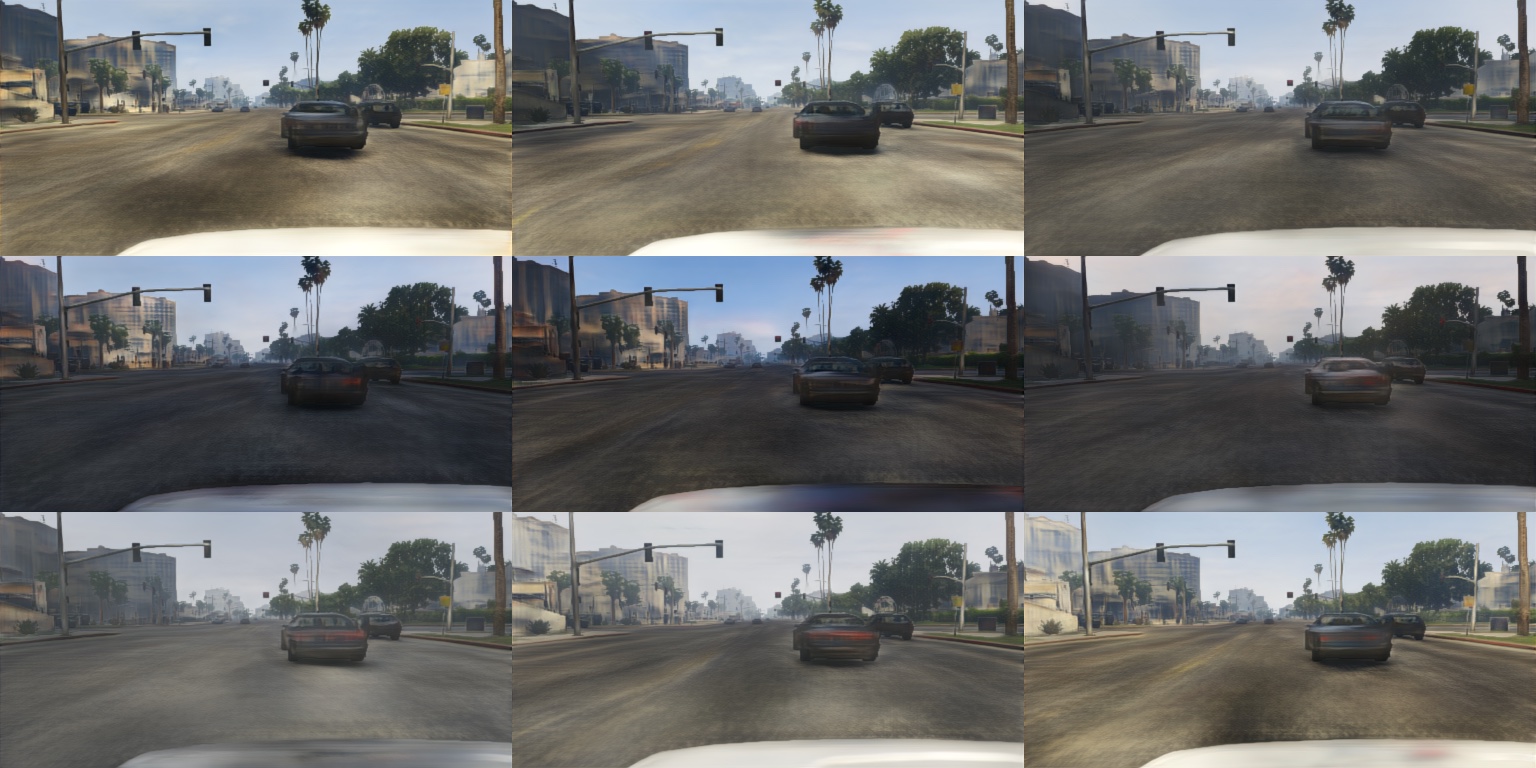}
    \caption{Our model w/o the rebalancing scheme}
    \end{subfigure}
    \begin{subfigure}{0.47\textwidth}
    \includegraphics[width=\linewidth]{fig/scene_2_ours.jpg}
    \caption{Our model }
    \end{subfigure}
    \caption{Ablation study using the same semantiic layout as Fig. \ref{fig:samp}.}
    \label{fig:ablation}
\end{figure*}
\begin{figure*}
    \centering
    \begin{subfigure}{\textwidth}
    \centering
    \includegraphics[width=0.9\linewidth]{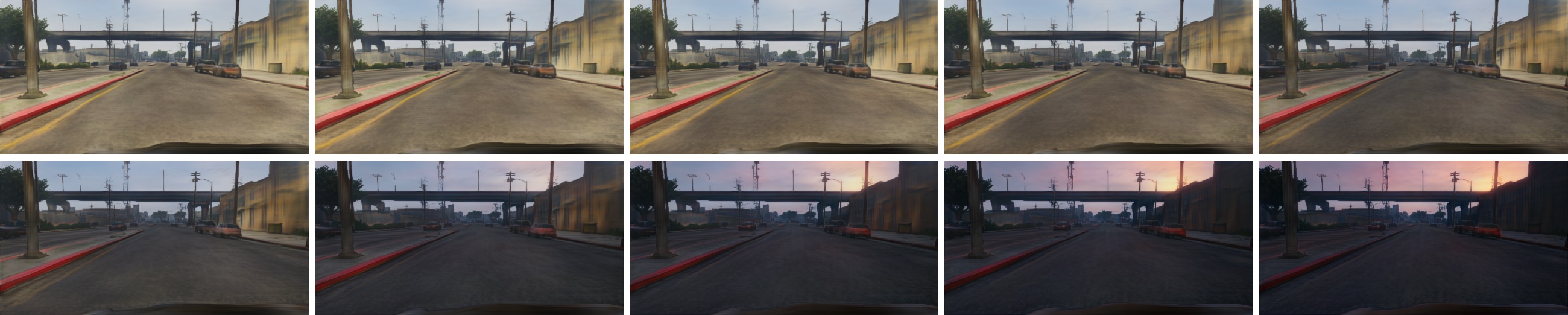}
    \caption{Change from daytime to night time}
    \end{subfigure}
    \begin{subfigure}{\textwidth}
    \centering
    \includegraphics[width=0.9\linewidth]{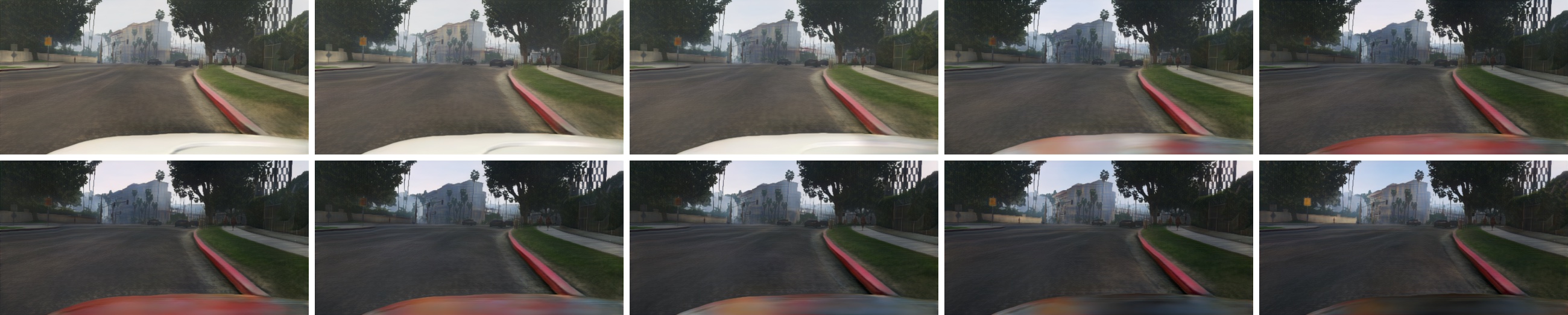}
    \caption{Change of car colours}
    \end{subfigure}
    \caption{Images generated by interpolating between latent noise vectors.}
    \label{fig:interp}
\end{figure*}

\begin{figure*}[htbp]
    \centering
    \begin{subfigure}{0.19\textwidth}
    \includegraphics[width=\linewidth]{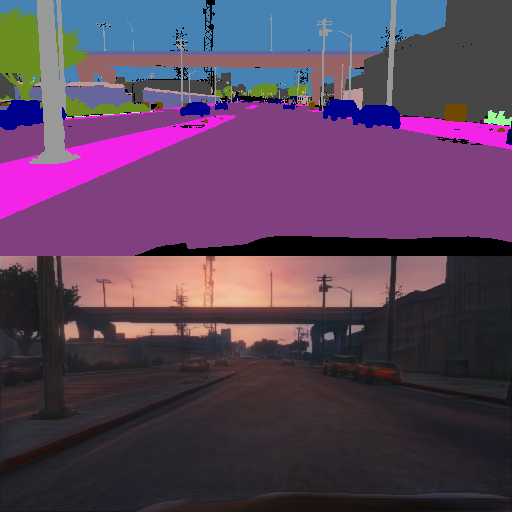}
    \caption{}
    \end{subfigure}
    \begin{subfigure}{0.19\textwidth}
    \includegraphics[width=\linewidth]{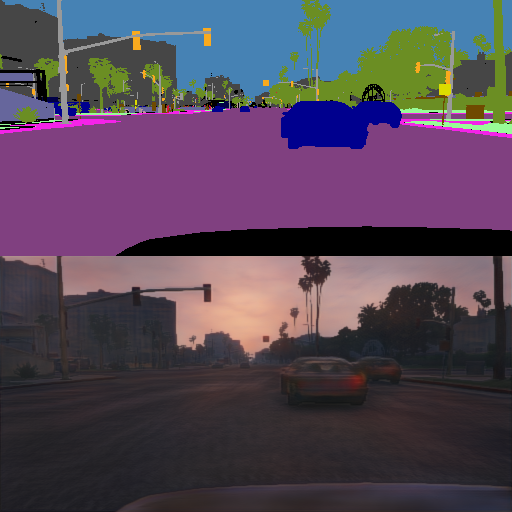}
    \caption{}
    \end{subfigure}
    \begin{subfigure}{0.19\textwidth}
    \includegraphics[width=\linewidth]{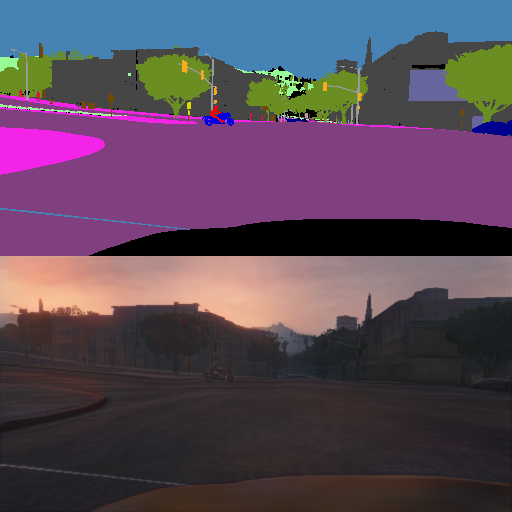}
    \caption{}
    \end{subfigure}
    \begin{subfigure}{0.19\textwidth}
    \includegraphics[width=\linewidth]{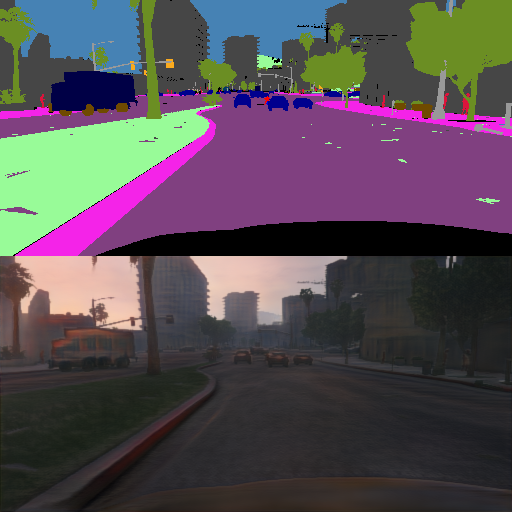}
    \caption{}
    \end{subfigure}
    \begin{subfigure}{0.19\textwidth}
    \includegraphics[width=\linewidth]{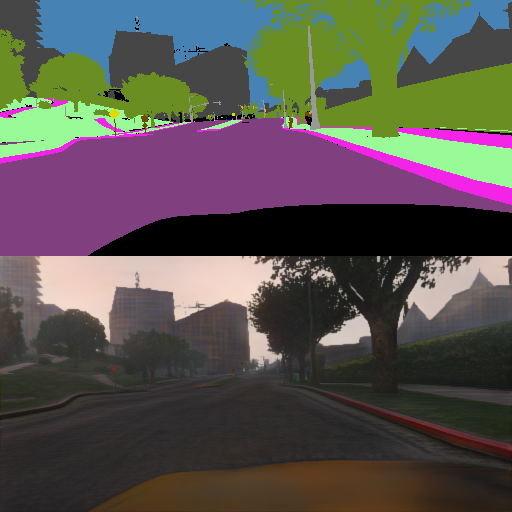}
    \caption{}
    \end{subfigure}
    \caption{Style consistency with the same random vector. (a) is the original input-output pair. We use the same random vector used in (a) and apply it to (b), (c), (d) and (e)}
    \label{fig:cons}
\end{figure*}

\subsubsection{Experimental Setting}

We train our model on 12403 training images and evaluate on the validation set (6383 images). Due to computational resource limitations, we conduct experiments at the $256 \times 512$ resolution. The noise encoder takes in an input noise $\widetilde{\mathbf{z}}$ of size $256 \times 512 \times 10$ and a segmentation map $\mathbf{x}$ of size $256 \times 512 \times 30$ and outputs an encoded noise $\mathbf{z}'$ of size $256 \times 512 \times10$. The number of output channels in the intermediate layers is $100$. We set the hyperparameters shown in Algorithm~\ref{alg:cimle} to the following values: $|S|=400$, $m=10$, $L=10000$, $|\widetilde{S}|=1$ and $\eta=10^{-5}$. 

\subsubsection{Quantitative Evaluation}

\paragraph{Diversity Evaluation} We measure the diversity of each method by generating 40 pairs of output images for each of 100 input semantic layouts from the test set. We then compute the average distance between each pair of output images for each given input semantic layout, which is then averaged over all input semantic layouts. The distance metric we use is LPIPS~\citep{zhang2018unreasonable}, which is designed to measure perceptual dissimilarity. The results are shown in Table \ref{tab:lpips}. As shown, the proposed method outperforms the baselines by a large margin. We also perform an ablation study and find that the proposed method performs better than variants that remove the noise encoder or the rebalancing scheme, which demonstrates the value of each component of our method. In Appendix~\ref{sec:baselines_rebalancing}, we also compare to baselines trained with the proposed rebalancing scheme and find that the baselines do not improve significantly even after incorporating the proposed rebalancing scheme. This suggests that the baselines are unable to take advantage of the rebalancing and demonstrates another inherent advantage of the proposed method: unlike the baselines that only encourage diversity in the generated images, the proposed method aims to learn the underlying data distribution and so is more sensitive to changes in the underlying distribution as a result of rebalancing. 

\paragraph{Image Quality Evaluation}
We now evaluate the generated image quality by human evaluation. Since it is difficult for humans to compare images with different styles, we selected the images that are closest to the ground truth image in $\ell_1$ distance among the images generated by CRN and our method. We then asked 62 human subjects to evaluate the images generated for 20 semantic layouts. For each semantic layout, they were asked to compare the image generated by CRN to the image generated by our method and judge which image exhibited more obvious synthetic patterns. The result is shown in Table \ref{tab:human}.

\begin{table}
    \centering
    \footnotesize
\begin{tabular}{cc}
\toprule
     Model& LPIPS score\\
     \midrule
     CRN&$0.11$\\
     CRN+noise&$0.12$\\
     Ours w/o noise encoder&$0.10$\\
     Ours w/o rebalancing scheme&$0.17$\\
     Ours&$\textbf{0.19}$\\
     \bottomrule
\end{tabular}
\caption{LPIPS score. We show the average perceptual distance of different models (including ablation study) and our proposed model gained the highest diversity.}
\label{tab:lpips}
\end{table}
\begin{table}
    \centering
    \footnotesize
\begin{tabular}{cc}
\toprule
     & \% of Images Containing More Artifacts \\
     \midrule
     CRN&$0.636\pm 0.242$\\
     Our method&$\textbf{0.364}\pm 0.242$\\
     \bottomrule
\end{tabular}
\caption{Average percentage of images that are judged by humans to exhibit more obvious synthetic patterns. Lower is better. }
\label{tab:human}
\end{table}

\begin{figure}[htbp]
    \centering
    \begin{subfigure}{0.2\textwidth}
    \includegraphics[width=\linewidth]{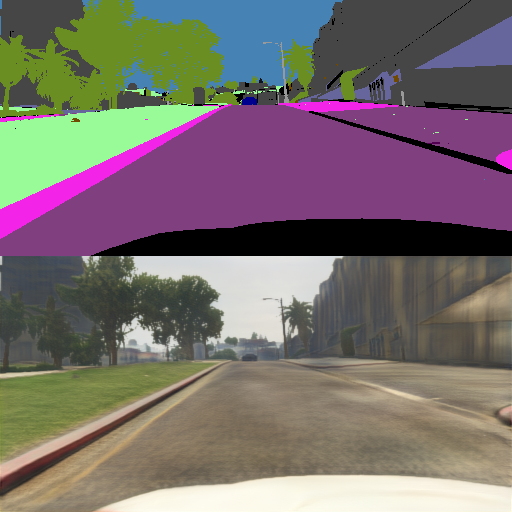}
    \caption{}
    \end{subfigure}
    \begin{subfigure}{0.2\textwidth}
    \includegraphics[width=\linewidth]{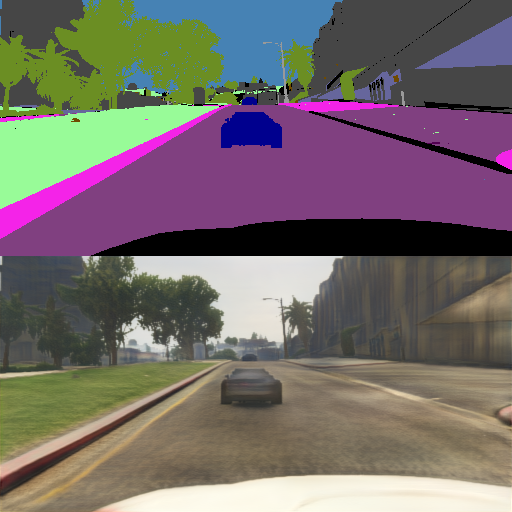}
    \caption{}
    \end{subfigure}
    \begin{subfigure}{0.2\textwidth}
    \includegraphics[width=\linewidth]{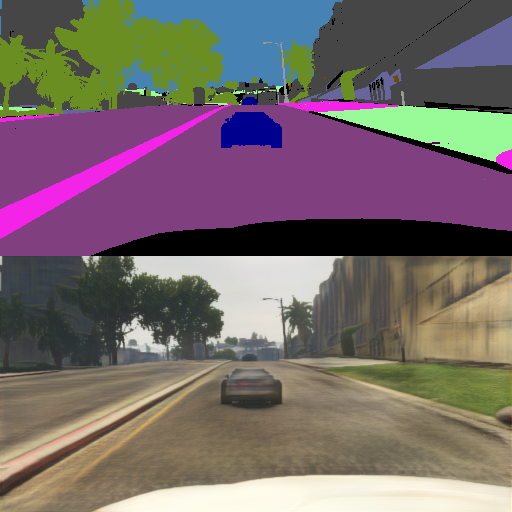}
    \caption{}
    \end{subfigure}
    \begin{subfigure}{0.2\textwidth}
    \includegraphics[width=\linewidth]{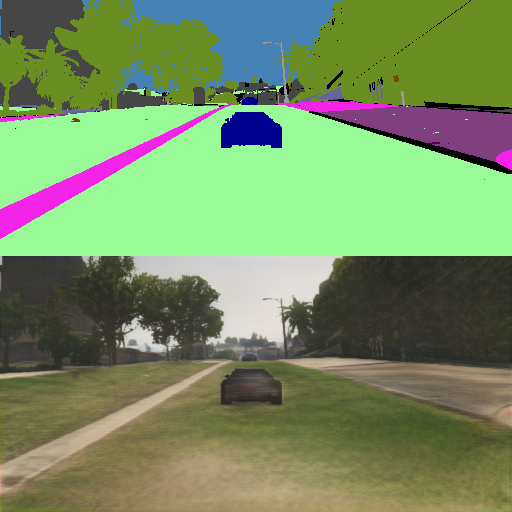}
    \caption{}
    \end{subfigure}
    \caption{Scene editing. (a) is the original input semantic map and the generated output. (b) adds a car on the road. (c) changes the grass on the left to road and change the side walk on the right to grass. (d) deletes our own car, changes the building on the right to tree and changes all road to grass.}
    \label{fig:edit}
\end{figure}

\subsubsection{Qualitative Evaluation}
A qualitative comparison is shown in Fig. \ref{fig:samp}. As shown, Pix2pix-HD generates almost identical images, BicycleGAN and MR-GAN generate images with heavy distortions, and CRN generates images with limited diversity. Here we tried to replace the generator of BicycleGAN~\citep{zhu2017toward} with the CRN~\citep{chen2017photographic} architecture to provide fair comparison, but training was very unstable, despite extensive hyperparameter tuning. (We reached out to the authors of BicycleGAN~\citep{zhu2017toward}, who confirmed that it is difficult to get BicycleGAN to work with the CRN architecture, and that to their knowledge, nobody has had success.) In comparison, the images generated by our method are diverse and do not suffer from distortions. We also perform an ablation study in Fig. \ref{fig:ablation}, which shows that each component of our method is important. Additionally, in Appendix~\ref{sec:baselines_rebalancing}, we compare to baselines trained with the proposed rebalancing scheme, which shows that the baselines are unable to take advantage of the rebalancing, unlike the proposed method. 

\paragraph{Interpolation}
We also perform linear interpolation of noise vectors to evaluate the quality of the learned latent space of noise vectors. As shown in \ref{fig:interp}(a), by interpolating between the noise vectors corresponding to generated images during daytime and nighttime respectively, we obtain a smooth transition from daytime to nighttime. We also show the transition in car colour in \ref{fig:interp}(b). This suggests that the learned latent space is sensible and captures the variation along both the daytime-nightime axis and the color axis. More examples and animations are available in Appendix~\ref{sec:additional_results}.

\paragraph{Scene Editing}
A successful method for image synthesis from semantic layouts enables users to manually edit the semantic map to synthesize desired imagery. One can do this simply by adding/deleting objects or changing the class label of a certain object. In Figure \ref{fig:edit} we show several such changes. Note that all four inputs use the same random vector; as shown, the images are highly consistent in terms of style, which is quite useful because the style should remain the same after editing the layout. We further demonstrate this in Fig. \ref{fig:cons} where we apply the random vector used in (a) to vastly different segmentation maps in (b),(c),(d),(e) and the sunset style is preserved across the different segmentation maps.

\section{Conclusion}

In this paper, we developed a new multimodal conditional image synthesis method by extending Implicit Maximum Likelihood Estimation (IMLE) to the conditional setting. We applied it to two different tasks: single image super-resolution and image synthesis from scene layout. On both tasks, we demonstrated that our method is able to generate multiple output images that are both different from each other and consistent with the input image even though only one ground truth output image per input image is provided as supervision. This compares favourably to the existing image synthesis methods, which either generate identical output images for the same input image due to mode collapse or generate spurious output images that are not consistent with the input image.

\paragraph{Acknowledgements.} This work was supported by ONR MURI N00014-14-1-0671. Ke Li thanks the Natural Sciences and Engineering Research Council of Canada (NSERC) for fellowship support. 


{\small
\bibliographystyle{spbasic}
\bibliography{cimle}

\begin{thebibliography}{55}
\providecommand{\natexlab}[1]{#1}
\providecommand{\url}[1]{{#1}}
\providecommand{\urlprefix}{URL }
\expandafter\ifx\csname urlstyle\endcsname\relax
  \providecommand{\doi}[1]{DOI~\discretionary{}{}{}#1}\else
  \providecommand{\doi}{DOI~\discretionary{}{}{}\begingroup
  \urlstyle{rm}\Url}\fi
\providecommand{\eprint}[2][]{\url{#2}}

\bibitem[{Almahairi et~al.(2018)Almahairi, Rajeswar, Sordoni, Bachman, and
  Courville}]{Almahairi2018AugmentedCL}
Almahairi A, Rajeswar S, Sordoni A, Bachman P, Courville A (2018) Augmented
  cyclegan: Learning many-to-many mappings from unpaired data. In: ICML

\bibitem[{Arjovsky and Bottou(2017)}]{arjovsky2017towards}
Arjovsky M, Bottou L (2017) Towards principled methods for training generative
  adversarial networks. arXiv preprint arXiv:170104862

\bibitem[{Arora and Zhang(2017)}]{arora2017gans}
Arora S, Zhang Y (2017) Do {GANs} actually learn the distribution? an empirical
  study. arXiv preprint arXiv:170608224

\bibitem[{Bruna et~al.(2015)Bruna, Sprechmann, and LeCun}]{bruna2015super}
Bruna J, Sprechmann P, LeCun Y (2015) Super-resolution with deep convolutional
  sufficient statistics. arXiv preprint arXiv:151105666

\bibitem[{Charpiat et~al.(2008)Charpiat, Hofmann, and
  Sch{\"o}lkopf}]{charpiat2008automatic}
Charpiat G, Hofmann M, Sch{\"o}lkopf B (2008) Automatic image colorization via
  multimodal predictions. In: European conference on computer vision, Springer,
  pp 126--139

\bibitem[{Chen and Koltun(2017)}]{chen2017photographic}
Chen Q, Koltun V (2017) Photographic image synthesis with cascaded refinement
  networks. In: IEEE International Conference on Computer Vision (ICCV), vol~1,
  p~3

\bibitem[{Cordts et~al.(2016)Cordts, Omran, Ramos, Rehfeld, Enzweiler,
  Benenson, Franke, Roth, and Schiele}]{Cordts2016Cityscapes}
Cordts M, Omran M, Ramos S, Rehfeld T, Enzweiler M, Benenson R, Franke U, Roth
  S, Schiele B (2016) The cityscapes dataset for semantic urban scene
  understanding. In: Proc. of the IEEE Conference on Computer Vision and
  Pattern Recognition (CVPR)

\bibitem[{Dahl et~al.(2017)Dahl, Norouzi, and Shlens}]{Dahl2017PixelRS}
Dahl R, Norouzi M, Shlens J (2017) Pixel recursive super resolution. 2017 IEEE
  International Conference on Computer Vision (ICCV) pp 5449--5458

\bibitem[{Denton et~al.(2015)Denton, Chintala, Fergus et~al.}]{denton2015deep}
Denton EL, Chintala S, Fergus R, et~al. (2015) Deep generative image models
  using a laplacian pyramid of adversarial networks. In: Advances in neural
  information processing systems, pp 1486--1494

\bibitem[{Donahue et~al.(2016)Donahue, Kr{\"a}henb{\"u}hl, and
  Darrell}]{donahue2016adversarial}
Donahue J, Kr{\"a}henb{\"u}hl P, Darrell T (2016) Adversarial feature learning.
  arXiv preprint arXiv:160509782

\bibitem[{Dumoulin et~al.(2016)Dumoulin, Belghazi, Poole, Mastropietro, Lamb,
  Arjovsky, and Courville}]{dumoulin2016adversarially}
Dumoulin V, Belghazi I, Poole B, Mastropietro O, Lamb A, Arjovsky M, Courville
  A (2016) Adversarially learned inference. arXiv preprint arXiv:160600704

\bibitem[{Finn et~al.(2016)Finn, Goodfellow, and Levine}]{finn2016unsupervised}
Finn C, Goodfellow I, Levine S (2016) Unsupervised learning for physical
  interaction through video prediction. In: Advances in neural information
  processing systems, pp 64--72

\bibitem[{Gauthier(2014)}]{gauthier2014conditional}
Gauthier J (2014) Conditional generative adversarial nets for convolutional
  face generation. Class Project for Stanford CS231N: Convolutional Neural
  Networks for Visual Recognition, Winter semester 2014(5):2

\bibitem[{Ghosh et~al.(2017)Ghosh, Kulharia, Namboodiri, Torr, and
  Dokania}]{ghosh2017multi}
Ghosh A, Kulharia V, Namboodiri V, Torr PH, Dokania PK (2017) Multi-agent
  diverse generative adversarial networks. arXiv preprint arXiv:170402906 1(4)

\bibitem[{Goodfellow et~al.(2014)Goodfellow, Pouget-Abadie, Mirza, Xu,
  Warde-Farley, Ozair, Courville, and Bengio}]{goodfellow2014generative}
Goodfellow I, Pouget-Abadie J, Mirza M, Xu B, Warde-Farley D, Ozair S,
  Courville A, Bengio Y (2014) Generative adversarial nets. In: Advances in
  neural information processing systems, pp 2672--2680

\bibitem[{Goodfellow(2014)}]{goodfellow2014distinguishability}
Goodfellow IJ (2014) On distinguishability criteria for estimating generative
  models. arXiv preprint arXiv:14126515

\bibitem[{Gutmann et~al.(2014)Gutmann, Dutta, Kaski, and
  Corander}]{gutmann2014likelihood}
Gutmann MU, Dutta R, Kaski S, Corander J (2014) Likelihood-free inference via
  classification. arXiv preprint arXiv:14074981

\bibitem[{Guzman-Rivera et~al.(2012)Guzman-Rivera, Batra, and
  Kohli}]{guzman2012multiple}
Guzman-Rivera A, Batra D, Kohli P (2012) Multiple choice learning: Learning to
  produce multiple structured outputs. In: Advances in Neural Information
  Processing Systems, pp 1799--1807

\bibitem[{Huang et~al.(2018)Huang, Liu, Belongie, and
  Kautz}]{huang2018multimodal}
Huang X, Liu MY, Belongie S, Kautz J (2018) Multimodal unsupervised
  image-to-image translation. arXiv preprint arXiv:180404732

\bibitem[{Isola et~al.(2017)Isola, Zhu, Zhou, and Efros}]{isola2017image}
Isola P, Zhu JY, Zhou T, Efros AA (2017) Image-to-image translation with
  conditional adversarial networks. arXiv preprint

\bibitem[{Johnson et~al.(2016)Johnson, Alahi, and
  Fei-Fei}]{johnson2016perceptual}
Johnson J, Alahi A, Fei-Fei L (2016) Perceptual losses for real-time style
  transfer and super-resolution. In: European Conference on Computer Vision,
  Springer, pp 694--711

\bibitem[{Kaneko et~al.(2017)Kaneko, Hiramatsu, and
  Kashino}]{kaneko2017generative}
Kaneko T, Hiramatsu K, Kashino K (2017) Generative attribute controller with
  conditional filtered generative adversarial networks. In: 2017 IEEE
  Conference on Computer Vision and Pattern Recognition (CVPR), IEEE, pp
  7006--7015

\bibitem[{Karacan et~al.(2016)Karacan, Akata, Erdem, and
  Erdem}]{karacan2016learning}
Karacan L, Akata Z, Erdem A, Erdem E (2016) Learning to generate images of
  outdoor scenes from attributes and semantic layouts. arXiv preprint
  arXiv:161200215

\bibitem[{Larsen et~al.(2015)Larsen, S{\o}nderby, Larochelle, and
  Winther}]{larsen2015autoencoding}
Larsen ABL, S{\o}nderby SK, Larochelle H, Winther O (2015) Autoencoding beyond
  pixels using a learned similarity metric. arXiv preprint arXiv:151209300

\bibitem[{Larsson et~al.(2016)Larsson, Maire, and
  Shakhnarovich}]{larsson2016learning}
Larsson G, Maire M, Shakhnarovich G (2016) Learning representations for
  automatic colorization. In: European Conference on Computer Vision, Springer,
  pp 577--593

\bibitem[{Ledig et~al.(2017)Ledig, Theis, Husz{\'a}r, Caballero, Cunningham,
  Acosta, Aitken, Tejani, Totz, Wang et~al.}]{ledig2017photo}
Ledig C, Theis L, Husz{\'a}r F, Caballero J, Cunningham A, Acosta A, Aitken AP,
  Tejani A, Totz J, Wang Z, et~al. (2017) Photo-realistic single image
  super-resolution using a generative adversarial network. In: CVPR, vol~2, p~4

\bibitem[{Lee et~al.(2018)Lee, Tseng, Mao, Huang, Lu, Singh, and
  Yang}]{Lee2018DRITDI}
Lee HY, Tseng HY, Mao Q, Huang JB, Lu YD, Singh MK, Yang MH (2018) Drit++:
  Diverse image-to-image translation via disentangled representations. ArXiv
  abs/1808.00948

\bibitem[{Lee et~al.(2019)Lee, Ha, and Kim}]{Lee2019HarmonizingML}
Lee S, Ha J, Kim G (2019) Harmonizing maximum likelihood with gans for
  multimodal conditional generation. ArXiv abs/1902.09225

\bibitem[{Li and Wand(2016)}]{li2016precomputed}
Li C, Wand M (2016) Precomputed real-time texture synthesis with markovian
  generative adversarial networks. In: European Conference on Computer Vision,
  Springer, pp 702--716

\bibitem[{Li and Malik(2016)}]{li2016fast}
Li K, Malik J (2016) Fast k-nearest neighbour search via {Dynamic Continuous
  Indexing}. In: International Conference on Machine Learning, pp 671--679

\bibitem[{Li and Malik(2017)}]{li2017fast}
Li K, Malik J (2017) Fast k-nearest neighbour search via {Prioritized DCI}. In:
  International Conference on Machine Learning, pp 2081--2090

\bibitem[{Li and Malik(2018)}]{li2018implicit}
Li K, Malik J (2018) Implicit maximum likelihood estimation. arXiv preprint
  arXiv:180909087

\bibitem[{Ma et~al.(2018)Ma, Jia, Georgoulis, Tuytelaars, and
  Gool}]{Ma2018ExemplarGU}
Ma L, Jia X, Georgoulis S, Tuytelaars T, Gool LV (2018) Exemplar guided
  unsupervised image-to-image translation with semantic consistency. In: ICLR

\bibitem[{Mathieu et~al.(2015)Mathieu, Couprie, and LeCun}]{mathieu2015deep}
Mathieu M, Couprie C, LeCun Y (2015) Deep multi-scale video prediction beyond
  mean square error. arXiv preprint arXiv:151105440

\bibitem[{Mirza and Osindero(2014)}]{mirza2014conditional}
Mirza M, Osindero S (2014) Conditional generative adversarial nets. arXiv
  preprint arXiv:14111784

\bibitem[{Oh et~al.(2015)Oh, Guo, Lee, Lewis, and Singh}]{oh2015action}
Oh J, Guo X, Lee H, Lewis RL, Singh S (2015) Action-conditional video
  prediction using deep networks in atari games. In: Advances in neural
  information processing systems, pp 2863--2871

\bibitem[{Pathak et~al.(2016)Pathak, Krahenbuhl, Donahue, Darrell, and
  Efros}]{pathak2016context}
Pathak D, Krahenbuhl P, Donahue J, Darrell T, Efros AA (2016) Context encoders:
  Feature learning by inpainting. In: Proceedings of the IEEE Conference on
  Computer Vision and Pattern Recognition, pp 2536--2544

\bibitem[{Reed et~al.(2016)Reed, Akata, Mohan, Tenka, Schiele, and
  Lee}]{reed2016learning}
Reed SE, Akata Z, Mohan S, Tenka S, Schiele B, Lee H (2016) Learning what and
  where to draw. In: Advances in Neural Information Processing Systems, pp
  217--225

\bibitem[{Richter et~al.(2016)Richter, Vineet, Roth, and
  Koltun}]{Richter_2016_ECCV}
Richter SR, Vineet V, Roth S, Koltun V (2016) Playing for data: {G}round truth
  from computer games. In: Leibe B, Matas J, Sebe N, Welling M (eds) European
  Conference on Computer Vision (ECCV), Springer International Publishing,
  LNCS, vol 9906, pp 102--118

\bibitem[{Sangkloy et~al.(2017)Sangkloy, Lu, Fang, Yu, and
  Hays}]{sangkloy2017scribbler}
Sangkloy P, Lu J, Fang C, Yu F, Hays J (2017) Scribbler: Controlling deep image
  synthesis with sketch and color. In: IEEE Conference on Computer Vision and
  Pattern Recognition (CVPR), vol~2

\bibitem[{Simonyan and Zisserman(2014)}]{simonyan2014very}
Simonyan K, Zisserman A (2014) Very deep convolutional networks for large-scale
  image recognition. arXiv preprint arXiv:14091556

\bibitem[{Srivastava et~al.(2015)Srivastava, Mansimov, and
  Salakhudinov}]{srivastava2015unsupervised}
Srivastava N, Mansimov E, Salakhudinov R (2015) Unsupervised learning of video
  representations using lstms. In: International conference on machine
  learning, pp 843--852

\bibitem[{Vondrick et~al.(2016)Vondrick, Pirsiavash, and
  Torralba}]{vondrick2016generating}
Vondrick C, Pirsiavash H, Torralba A (2016) Generating videos with scene
  dynamics. In: Advances In Neural Information Processing Systems, pp 613--621

\bibitem[{Wang et~al.(2017)Wang, Liu, Zhu, Tao, Kautz, and
  Catanzaro}]{wang2017high}
Wang TC, Liu MY, Zhu JY, Tao A, Kautz J, Catanzaro B (2017) High-resolution
  image synthesis and semantic manipulation with conditional gans. arXiv
  preprint arXiv:171111585

\bibitem[{Wang and Gupta(2016)}]{wang2016generative}
Wang X, Gupta A (2016) Generative image modeling using style and structure
  adversarial networks. In: European Conference on Computer Vision, Springer,
  pp 318--335

\bibitem[{Wang et~al.(2018)Wang, Yu, Wu, Gu, Liu, Dong, Loy, Qiao, and
  Tang}]{Wang2018ESRGANES}
Wang X, Yu K, Wu S, Gu J, Liu Y, Dong C, Loy CC, Qiao Y, Tang X (2018) Esrgan:
  Enhanced super-resolution generative adversarial networks. CoRR
  abs/1809.00219

\bibitem[{Yang et~al.(2019)Yang, Hong, Jang, Zhao, and
  Lee}]{Yang2019DiversitySensitiveCG}
Yang D, Hong S, Jang Y, Zhao T, Lee H (2019) Diversity-sensitive conditional
  generative adversarial networks. ArXiv abs/1901.09024

\bibitem[{Yoo et~al.(2016)Yoo, Kim, Park, Paek, and Kweon}]{yoo2016pixel}
Yoo D, Kim N, Park S, Paek AS, Kweon IS (2016) Pixel-level domain transfer. In:
  European Conference on Computer Vision, Springer, pp 517--532

\bibitem[{Yu et~al.(2018)Yu, Xian, Chen, Liu, Liao, Madhavan, and
  Darrell}]{yu2018bdd100k}
Yu F, Xian W, Chen Y, Liu F, Liao M, Madhavan V, Darrell T (2018) Bdd100k: A
  diverse driving video database with scalable annotation tooling. arXiv
  preprint arXiv:180504687

\bibitem[{Zhang et~al.(2016)Zhang, Isola, and Efros}]{zhang2016colorful}
Zhang R, Isola P, Efros AA (2016) Colorful image colorization. In: European
  Conference on Computer Vision, Springer, pp 649--666

\bibitem[{Zhang et~al.(2018)Zhang, Isola, Efros, Shechtman, and
  Wang}]{zhang2018unreasonable}
Zhang R, Isola P, Efros AA, Shechtman E, Wang O (2018) The unreasonable
  effectiveness of deep features as a perceptual metric. arXiv preprint

\bibitem[{Zhu et~al.(2017{\natexlab{a}})Zhu, Zhang, Pathak, Darrell, Efros,
  Wang, and Shechtman}]{DBLP:journals/corr/abs-1711-11586}
Zhu J, Zhang R, Pathak D, Darrell T, Efros AA, Wang O, Shechtman E
  (2017{\natexlab{a}}) Toward multimodal image-to-image translation. CoRR
  abs/1711.11586, \urlprefix\url{http://arxiv.org/abs/1711.11586},
  \eprint{1711.11586}

\bibitem[{Zhu et~al.(2016)Zhu, Kr{\"a}henb{\"u}hl, Shechtman, and
  Efros}]{zhu2016generative}
Zhu JY, Kr{\"a}henb{\"u}hl P, Shechtman E, Efros AA (2016) Generative visual
  manipulation on the natural image manifold. In: European Conference on
  Computer Vision, Springer, pp 597--613

\bibitem[{Zhu et~al.(2017{\natexlab{b}})Zhu, Park, Isola, and
  Efros}]{zhu2017unpaired}
Zhu JY, Park T, Isola P, Efros AA (2017{\natexlab{b}}) Unpaired image-to-image
  translation using cycle-consistent adversarial networks. arXiv preprint

\bibitem[{Zhu et~al.(2017{\natexlab{c}})Zhu, Zhang, Pathak, Darrell, Efros,
  Wang, and Shechtman}]{zhu2017toward}
Zhu JY, Zhang R, Pathak D, Darrell T, Efros AA, Wang O, Shechtman E
  (2017{\natexlab{c}}) Toward multimodal image-to-image translation. In:
  Advances in Neural Information Processing Systems, pp 465--476

\end{thebibliography}
}

\begin{appendices}

\newpage
\section{Baselines with Proposed Rebalancing Scheme}
\label{sec:baselines_rebalancing}

\begin{figure}[h]
    \centering
    \begin{subfigure}{0.45\textwidth}
    \includegraphics[width=\linewidth]{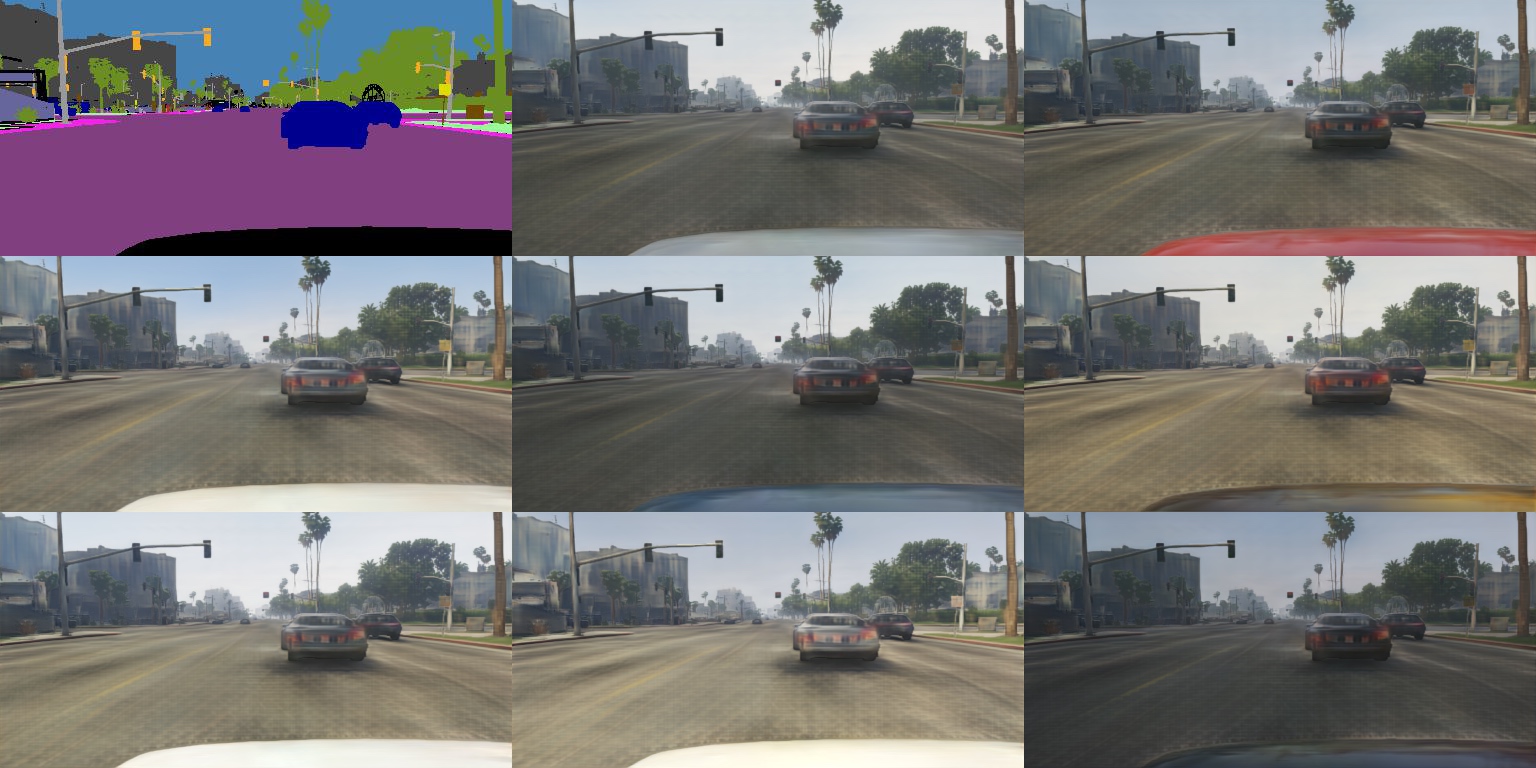}
    \caption{CRN}
    \end{subfigure}
    \begin{subfigure}{0.45\textwidth}
    \includegraphics[width=\linewidth]{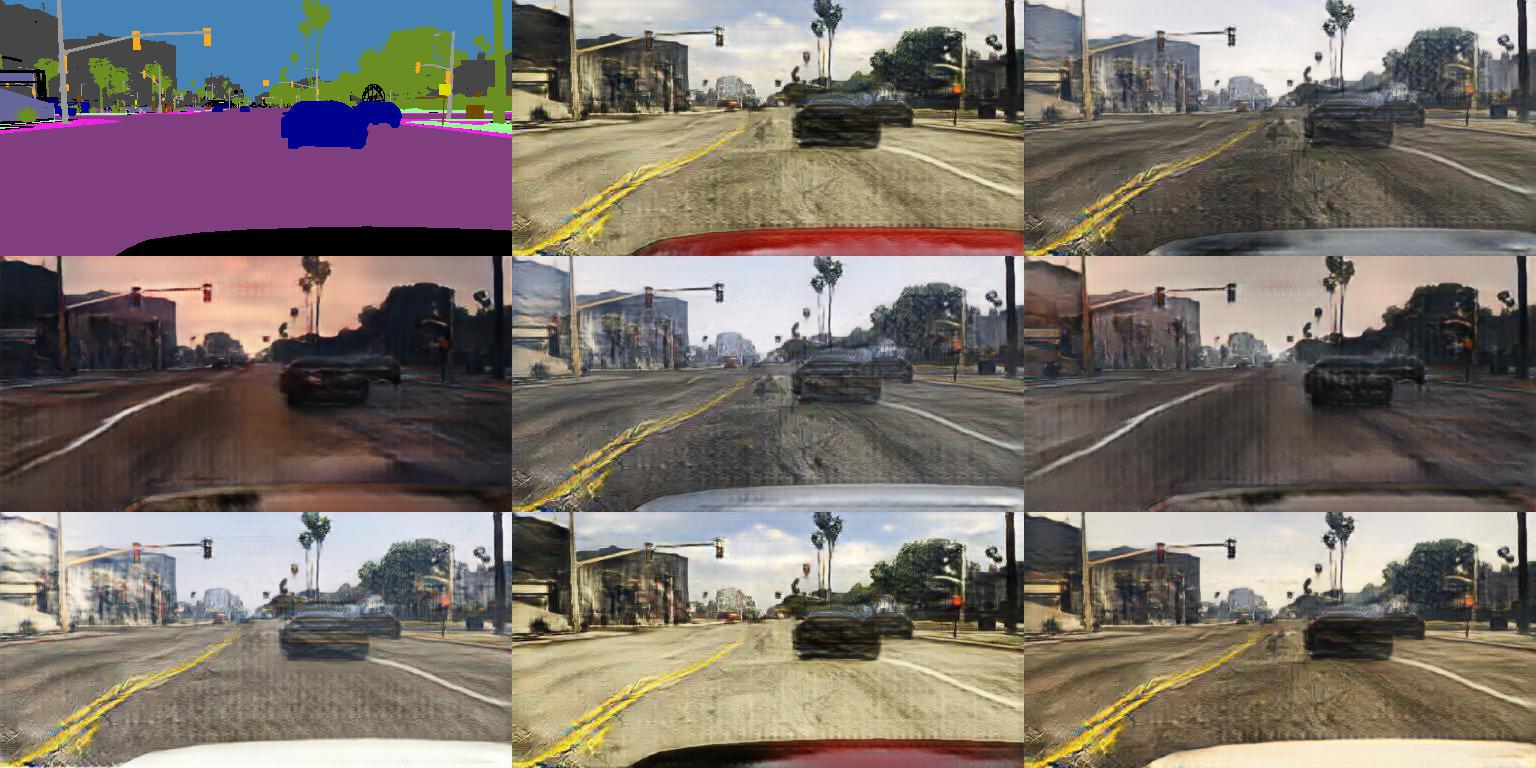}
    \caption{BicycleGAN}
    \end{subfigure}
    \begin{subfigure}{0.45\textwidth}
    \includegraphics[width=\linewidth]{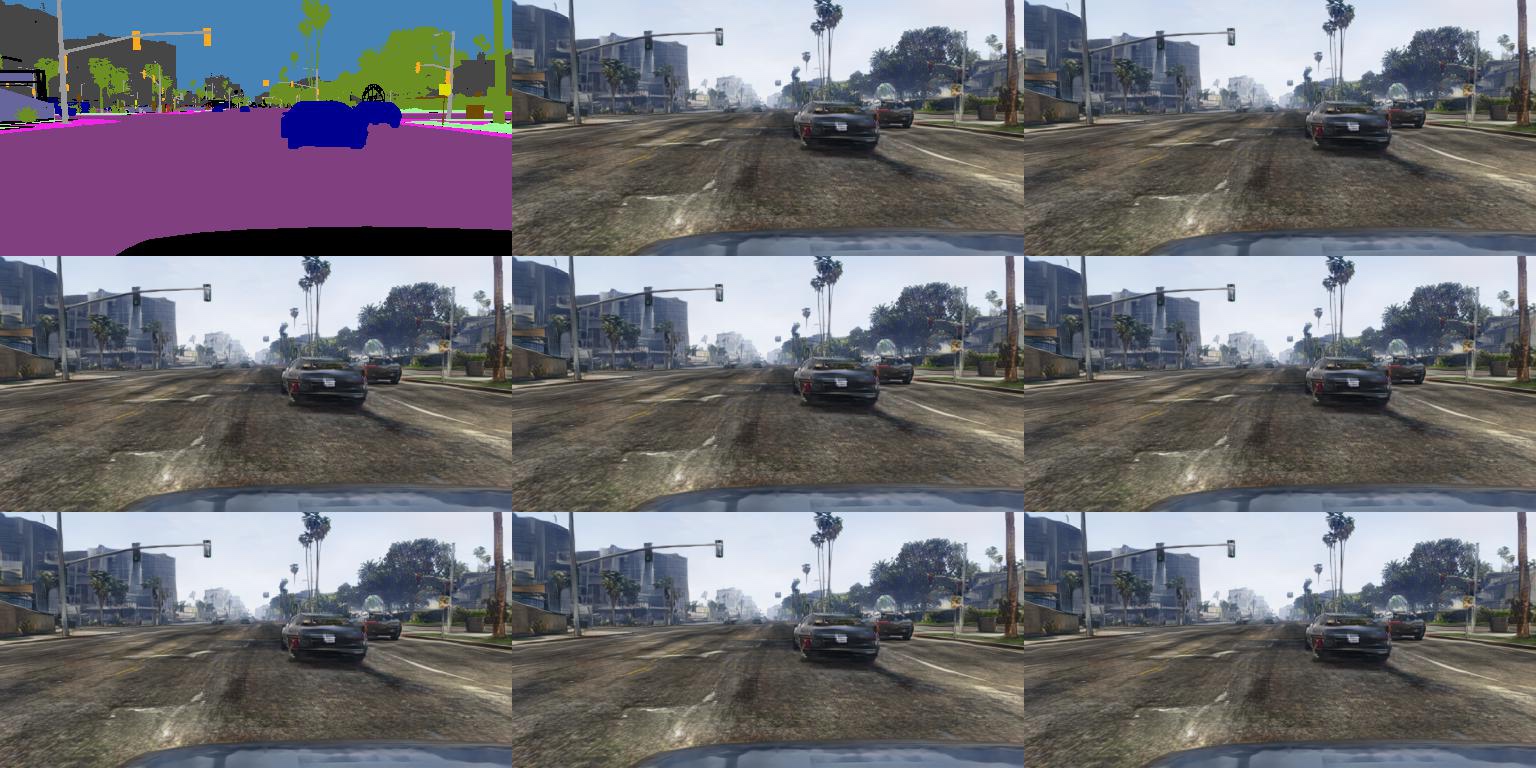}
    \caption{Pix2pix-HD}
    \end{subfigure}
    \begin{subfigure}{0.45\textwidth}
    \includegraphics[width=\linewidth]{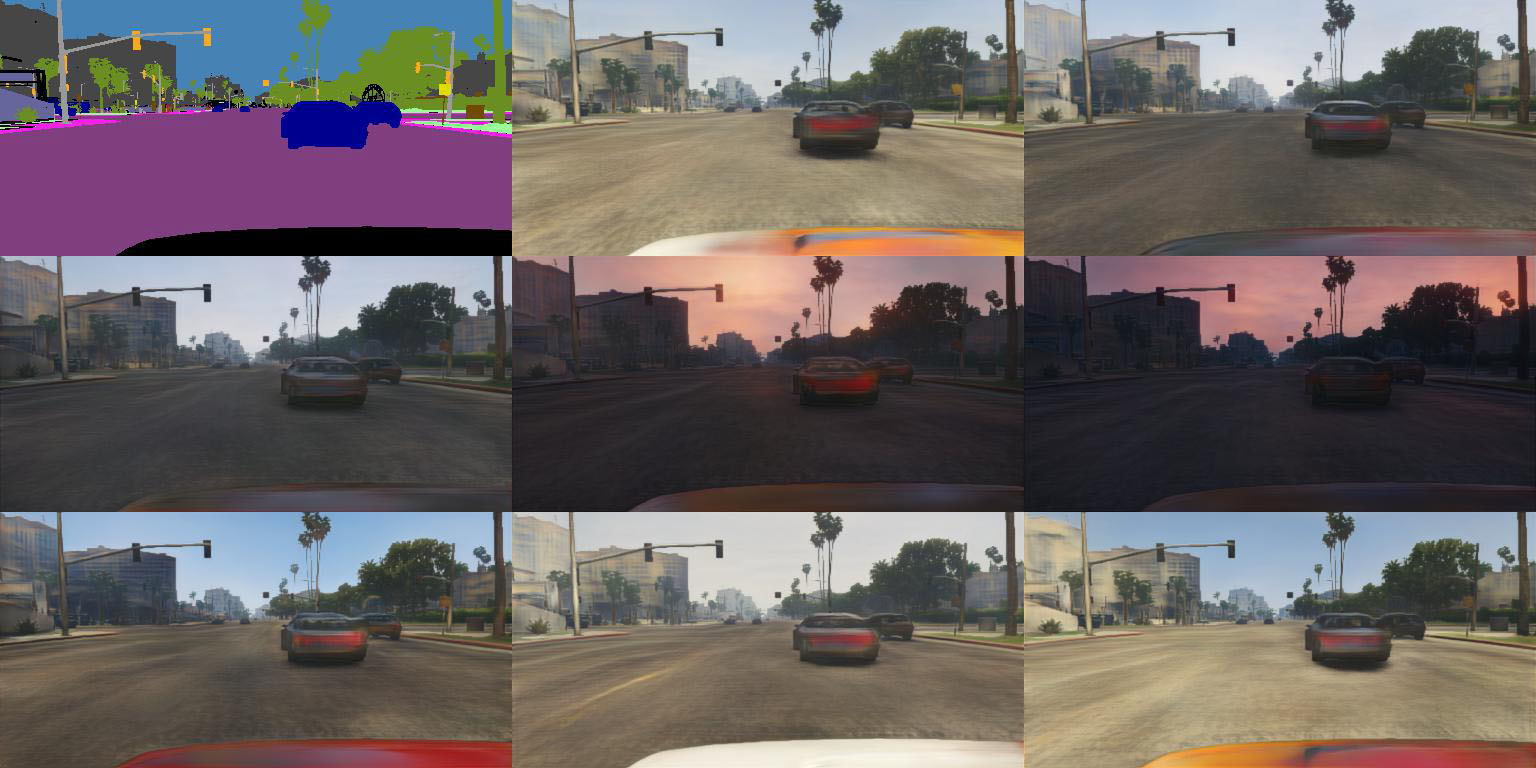}
    \caption{Ours}
    \end{subfigure}
    \caption{Samples generated by baselines with proposed rebalancing scheme, compared to the samples generated by our method. As shown, even with the proposed rebalancing scheme, the samples generated by CRN and Pix2pix-HD exhibit far less diversity than the samples generated by our method, and the samples generated by BicycleGAN are both less diverse and contain more artifacts than the samples generated by our method.}
    \label{fig:rebalance}
\end{figure}

A natural question is whether applying the proposed rebalancing scheme to the baselines would result in a significant improvement in the diversity of generated images. We tried this and found that the diversity is still lacking; the results are shown in Figure~\ref{fig:rebalance}. The LPIPS score of CRN only improves slightly from 0.12 to 0.13 after dataset and loss rebalancing are applied. It still underperforms our method, which achieves a LPIPS score of 0.19. The LPIPS score of Pix2pix-HD showed no improvement after applying dataset rebalancing; it still ignores the latent input noise vector. This suggests that the baselines are not able to take advantage of the rebalancing scheme. On the other hand, our method is able to take advantage of it, demonstrating its superior capability compared to the baselines. 

\newpage

\section{Additional Results}
\label{sec:additional_results}

All videos that we refer to below are available at \url{http://people.eecs.berkeley.edu/~ke.li/projects/imle/scene_layouts}.

\subsection{Video of Interpolations}

\begin{figure*}[h]
    \centering
    \begin{subfigure}{\textwidth}
    \includegraphics[width=\linewidth]{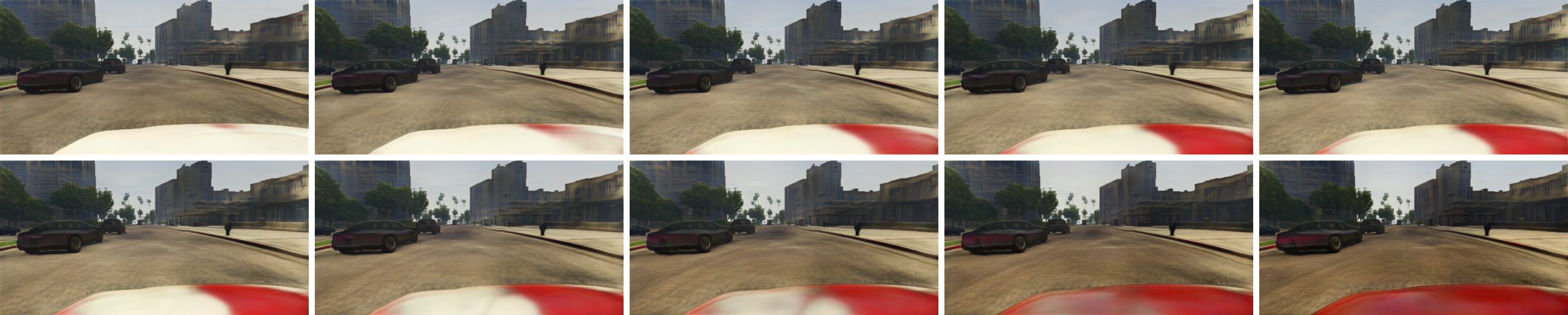}
    \caption{}
    \end{subfigure}
    \begin{subfigure}{\textwidth}
    \includegraphics[width=\linewidth]{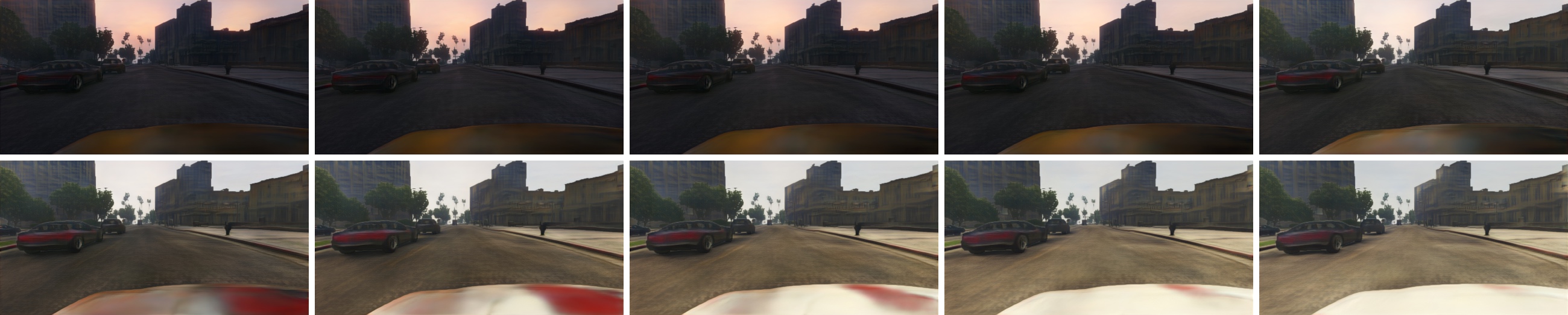}
    \caption{}
    \end{subfigure}
    \caption{\label{fig:interpvid}Frames of video generated by smoothly interpolating between latent noise vectors. }
    
\end{figure*}

We generated a video that shows smooth transitions between different renderings of the same scene. Frames of the generated video are shown in Figure~\ref{fig:interpvid}. 

\newpage
\subsection{Video Generation from Evolving Scene Layouts}

\begin{figure*}[h]
    \centering
    \begin{subfigure}{\textwidth}
    \includegraphics[width=\linewidth]{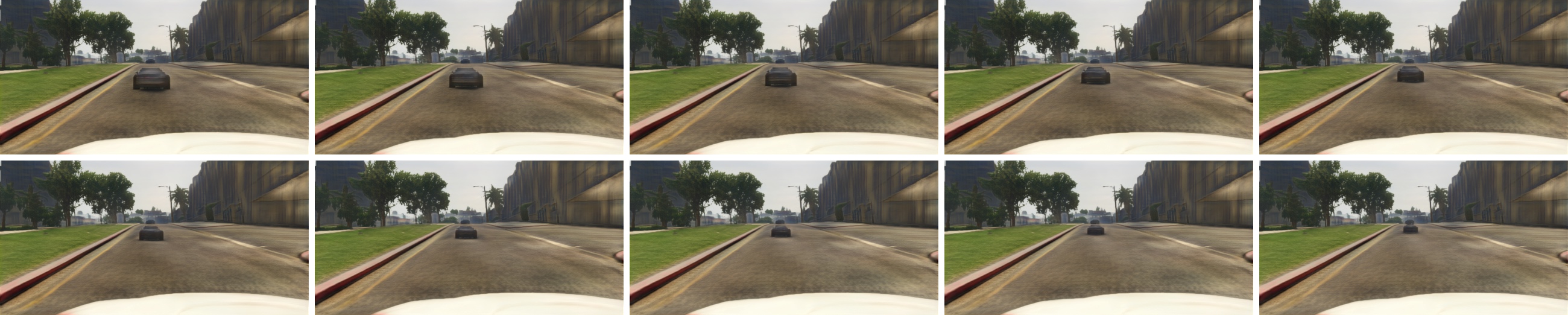}
    \caption{}
    \end{subfigure}
    \begin{subfigure}{\textwidth}
    \includegraphics[width=\linewidth]{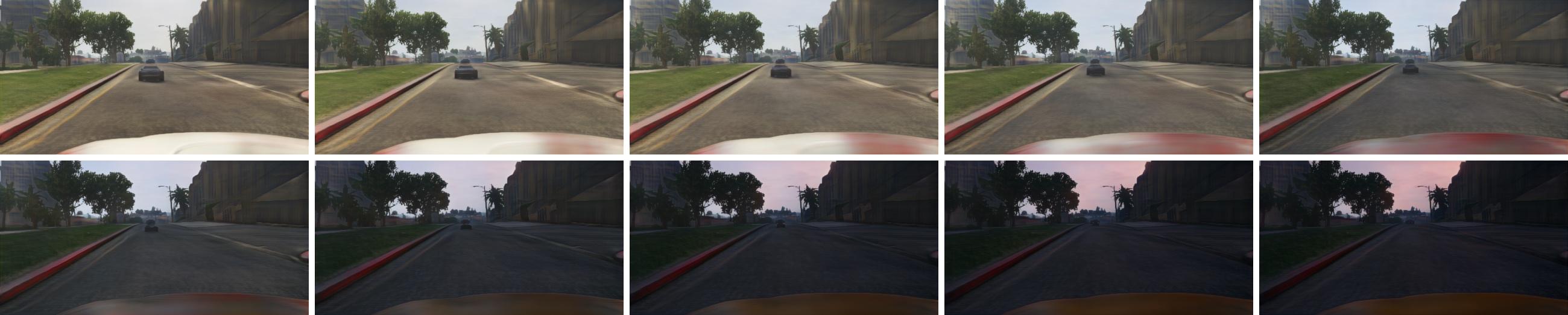}
    \caption{}
    \end{subfigure}
    \caption{\label{fig:vidgen}Frames from two videos of a moving car generated using our method. In both videos, we feed in scene layouts with cars of varying sizes to our model to generate different frames. In (a), we use the same latent noise vector across all frames. In (b), we interpolate between two latent noise vectors, one of which corresponds to a daytime scene and the other to a night time scene. The consistency of style across frames demonstrates that the learned space of latent noise vectors is semantically meaningful and that scene layout and style are successfully disentangled by our model. }
    
\end{figure*}

We generated videos of a car moving farther away from the camera and then back towards the camera by generating individual frames independently using our model with different semantic segmentation maps as input. For the video to have consistent appearance, we must be able to consistently select the same mode across all frames. In Figure~\ref{fig:vidgen}, we show that our model has this capability: we are able to select a mode consistently by using the same latent noise vector across all frames. 

\newpage

\begin{figure*}[h]
    \centering
    \begin{subfigure}{\textwidth}
    \includegraphics[width=\linewidth]{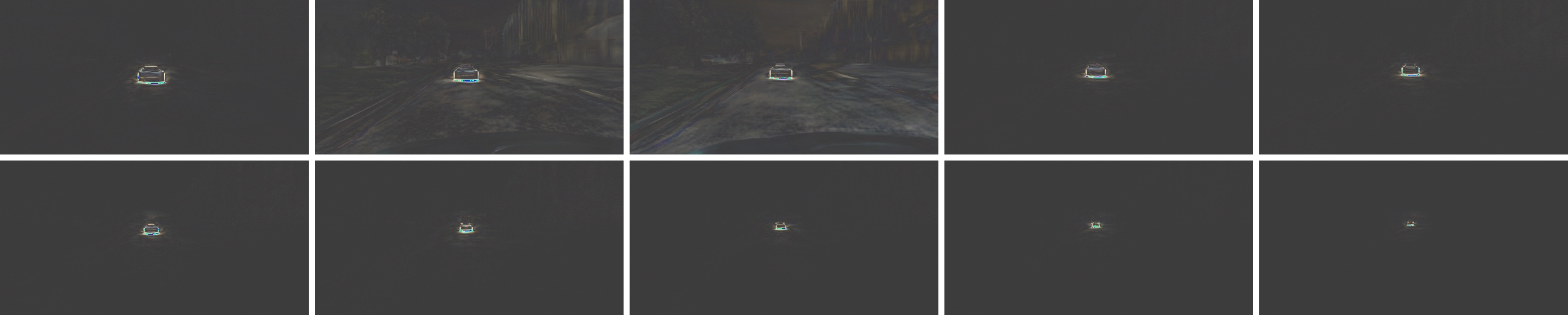}
    \caption{}
    \end{subfigure}
    \begin{subfigure}{\textwidth}
    \includegraphics[width=\linewidth]{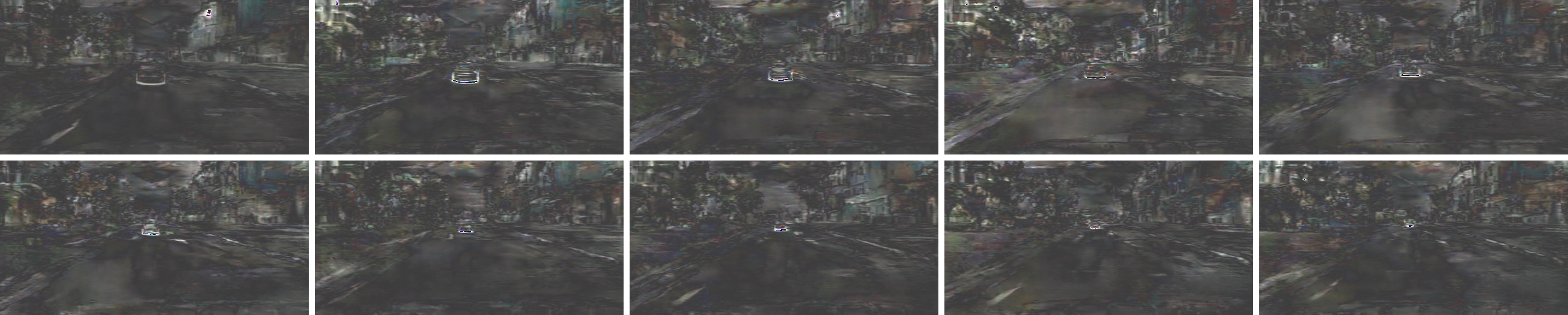}
    \caption{}
    \end{subfigure}
    \caption{\label{fig:vidgendiff}Comparison of the difference between adjacent frames of synthesized moving car video. Darker pixels indicate smaller difference and lighter pixels indicate larger difference. (a) shows results for the video generated by our model. (b) shows results for the video generated by pix2pix~\citep{isola2017image}. }
    
\end{figure*}

Here we demonstrate one potential benefit of modelling multiple modes instead of a single mode. We tried generating a video from the same sequence of scene layouts using pix2pix~\citep{isola2017image}, which only models a single mode. (For pix2pix, we used a pretrained model trained on Cityscapes, which is easier for the purposes of generating consistent frames because Cityscapes is less diverse than GTA-5.) In Figure~\ref{fig:vidgendiff}, we show the difference between adjacent frames in the videos generated by our model and pix2pix. As shown, our model is able to generate consistent appearance across frames (as evidenced by the small difference between adjacent frames). On the other hand, pix2pix is not able to generate consistent appearance across frames, because it arbitrarily picks a mode to generate and does not permit control over which mode it generates. 

\newpage
\section{More Generated Samples}

\begin{figure*}[ht]
    \centering
    \begin{subfigure}{\textwidth}
    \includegraphics[width=\linewidth]{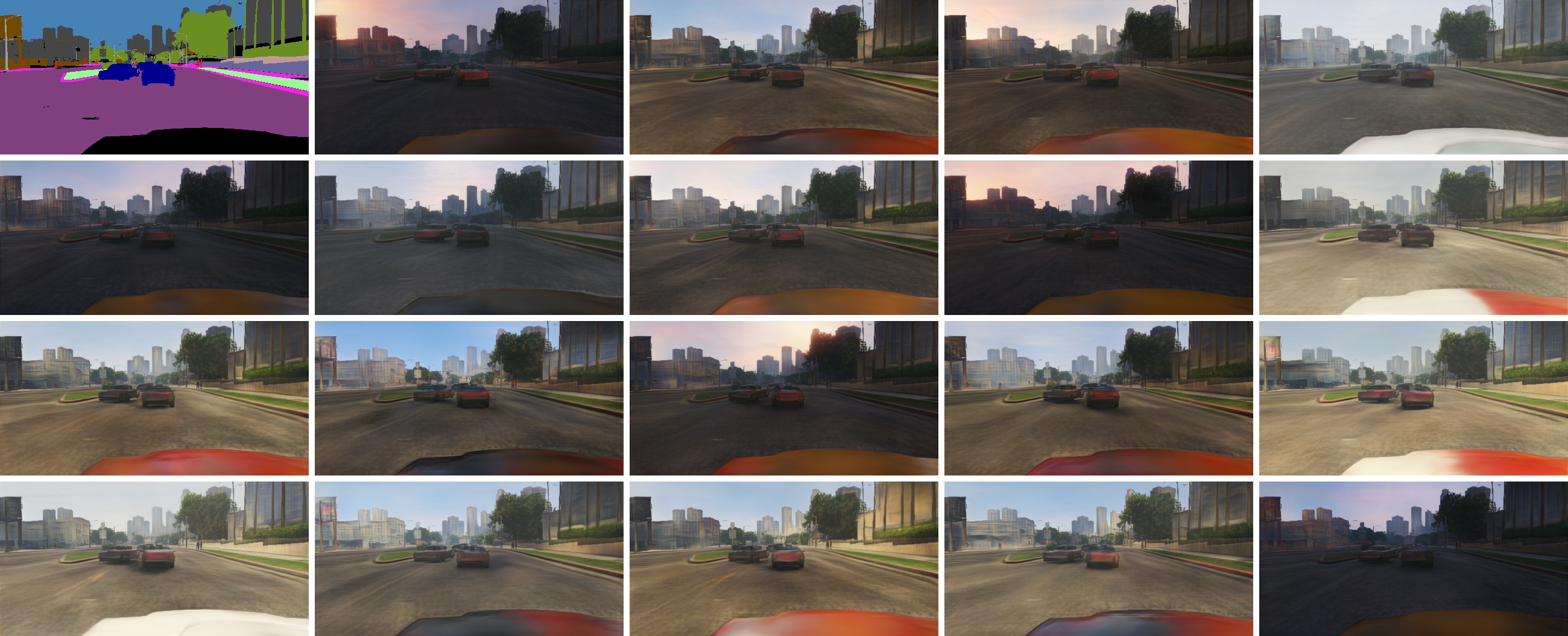}
    \caption{}
    \end{subfigure}
    \begin{subfigure}{\textwidth}
    \includegraphics[width=\linewidth]{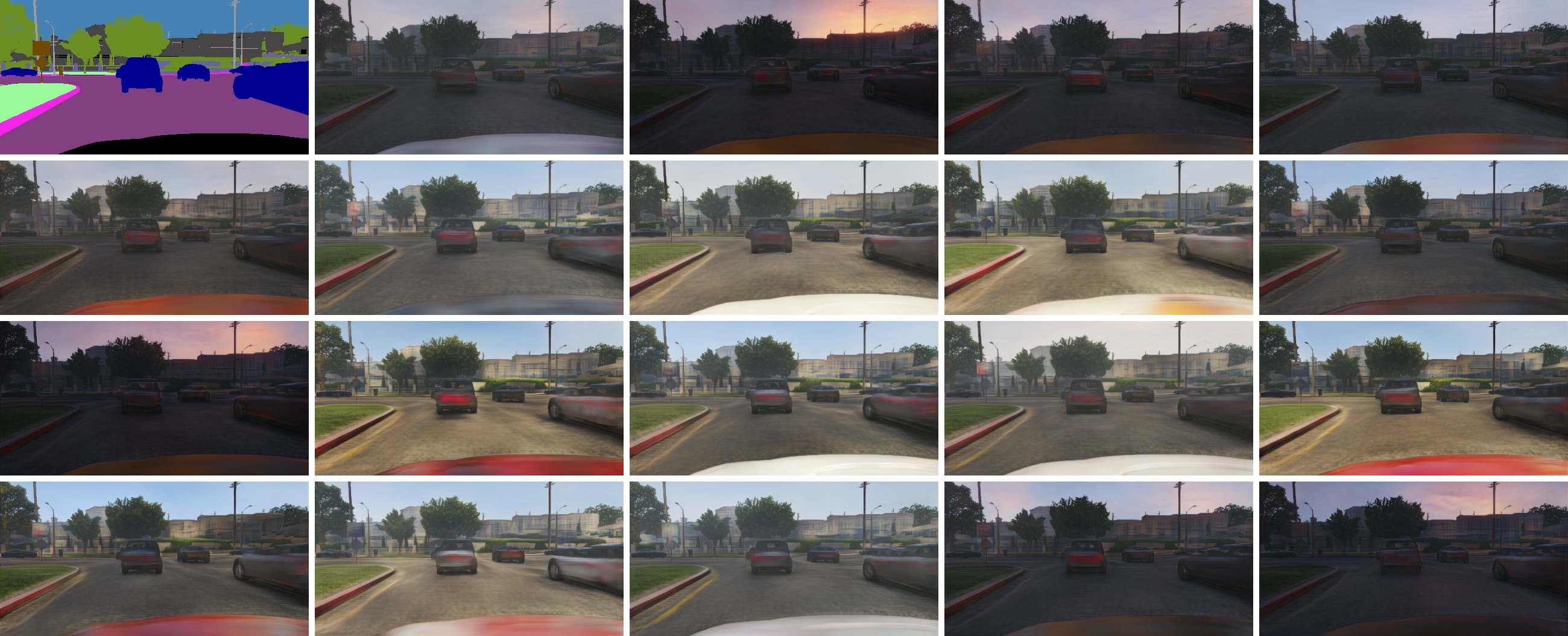}
    \caption{}
    \end{subfigure}
    \caption{Samples generated by our model. The image at the top-left corner is the input semantic layout and the other 19 images are samples generated by our model conditioned on the same semantic layout.}
    \label{fig:eg20_more}
\end{figure*}
\newpage
\begin{figure*}[ht]
    \centering
    \begin{subfigure}{\textwidth}
    \includegraphics[width=\linewidth]{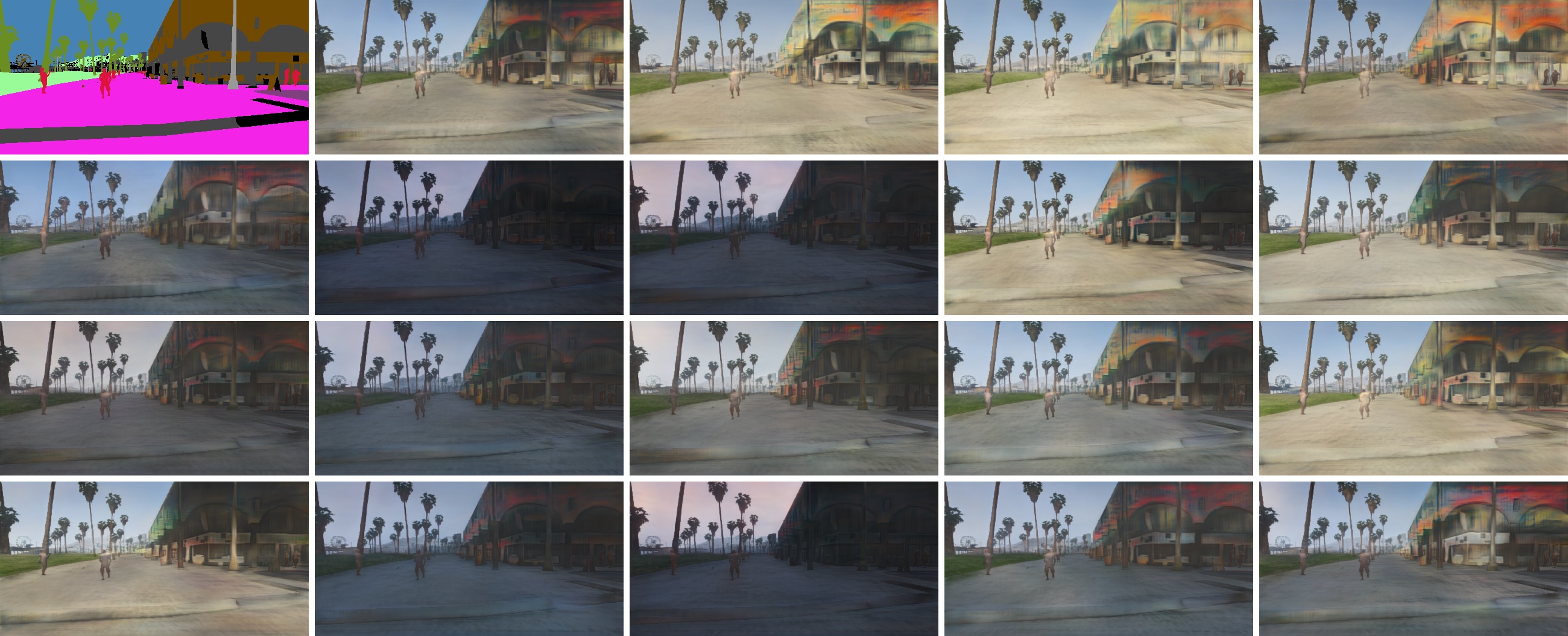}
    \caption{}
    \end{subfigure}
    \begin{subfigure}{\textwidth}
    \includegraphics[width=\linewidth]{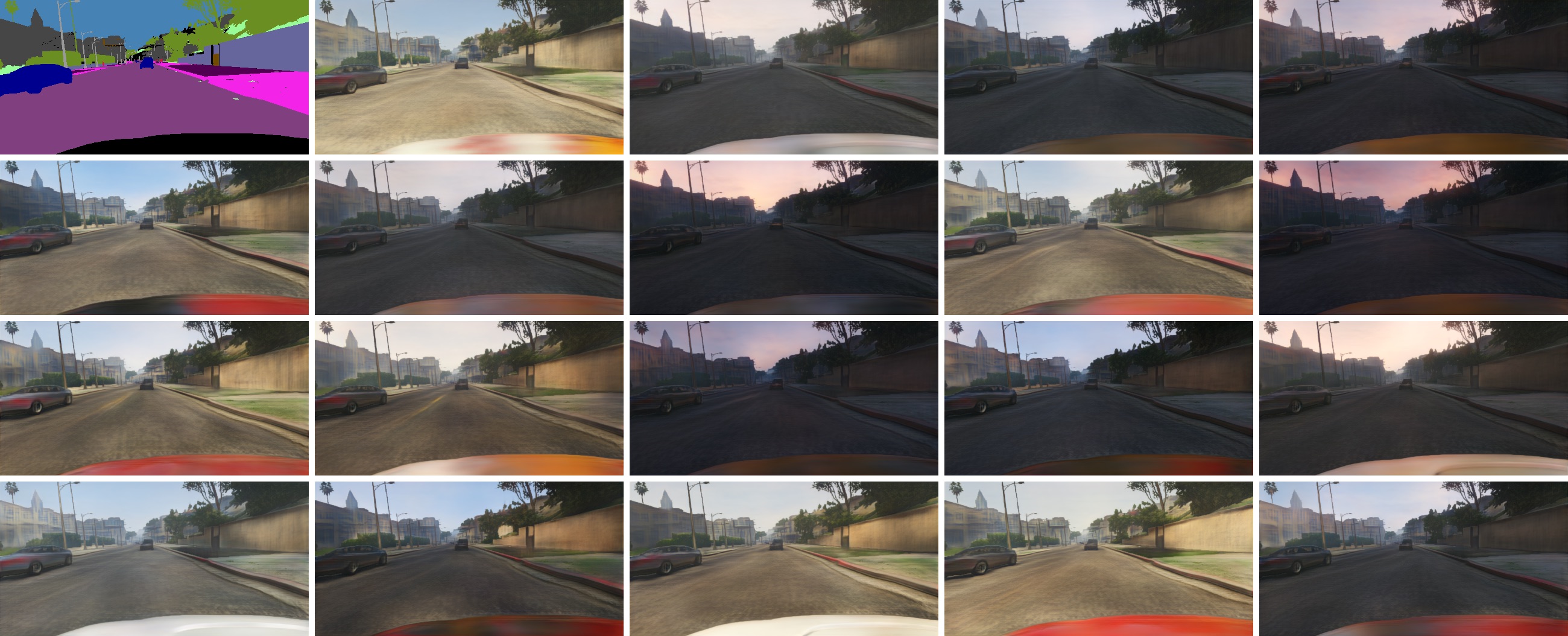}
    \caption{}
    \end{subfigure}
    \caption{Samples generated by our model. The image at the top-left corner is the input semantic layout and the other 19 images are samples generated by our model conditioned on the same semantic layout.}
    \label{fig:eg20_more2}
\end{figure*}

\end{appendices}

\end{document}